\documentclass{article}

\usepackage[preprint]{neurips_2026}

\usepackage[utf8]{inputenc} 
\usepackage[T1]{fontenc}    
\usepackage[para]{footmisc}
\usepackage{hyperref}       
\usepackage{url}            
\usepackage{booktabs}       
\usepackage{amsfonts}       
\usepackage{amsmath}
\usepackage{nicefrac}       
\usepackage{microtype}      
\usepackage{subcaption}
\usepackage{cleveref}
\usepackage{graphicx}
\usepackage{multirow}
\usepackage{multicol}
\usepackage{makecell}
\usepackage{pifont}
\usepackage{wrapfig}
\usepackage[dvipsnames]{xcolor}         
\usepackage[dvipsnames, table]{xcolor}
\usepackage{xspace}
\usepackage{colortbl}
\usepackage{dsfont}
\usepackage{listings}
\usepackage{wrapfig}
\usepackage[export]{adjustbox}
\usepackage{doi}
\newcommand{\modelName}{DriveJudge}
\newcommand{\method}{DriveJudge}

\newcommand{\cmark}{\ding{51}}%
\newcommand{\xmark}{\ding{55}}%

\definecolor{mygreen}{RGB}{220, 255, 220}
\definecolor{lgreen}{RGB}{236, 255, 201}
\definecolor{nvgreen}{RGB}{118, 185, 0}
\definecolor{lgreen}{RGB}{236, 255, 201}
\definecolor{nvgreen}{RGB}{118, 185, 0}
\definecolor{cvprblue}{rgb}{0.21,0.49,0.74}
\definecolor{lgray}{RGB}{245,245,245}

\definecolor{lightblue}{rgb}{0.678, 0.847, 0.902}

\title{DriveJudge: Rethinking Autonomous Driving Evaluation with Vision-Language Models}

\author{
Xinglong Sun\thanks{Primary contact: \texttt{xinglongs@nvidia.com}} \And Kevin Xie \And Jenny Schmalfuss \And Despoina Paschalidou \And Xiuming Zhang \And Sanja Fidler \And Kashyap Chitta\thanks{Work done at NVIDIA, currently affiliated with KE:SAI.} \And Jose M. Alvarez \AND
{
\large
NVIDIA
}
}
\begin{document}

\maketitle

\begin{figure}[h]
    \centering
    \vspace{-25pt}
    \includegraphics[width=\textwidth]{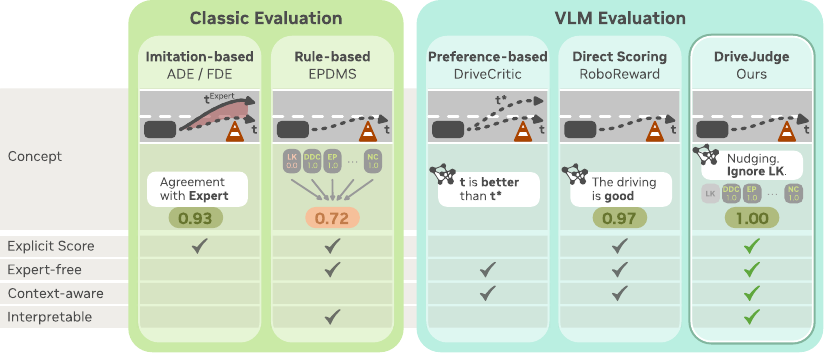}
    \caption{\textbf{Comparison of evaluation paradigms.} 
    We address limitations of existing classical and VLM-based metrics with \method{} which is a VLM-guided, rule-grounded evaluation paradigm. It selectively invokes rules based on scene context, enabling explicit, expert-free, context-aware, and interpretable driving assessment. In the example, \method{} ignores lane keeping (LK) to nudge.
    }
    \label{fig:teaser}
    \vspace{-5pt}
\end{figure}

\begin{abstract}
     Autonomous driving has shifted towards end-to-end policy learning, where reliable, interpretable policy evaluation is a fundamental challenge as driving quality is highly context-dependent.
     Commonly used rule-based driving metrics like EPDMS~\cite{dauner2024navsim} are interpretable but lack context-awareness, while recent VLM-based evaluations are context-aware but limited by ambiguous VLM outputs and weak physical grounding.
     To evaluate driving in a manner that is both interpretable and context-aware, we introduce \textbf{DriveJudge}.
     DriveJudge is a driving evaluation agent that combines rule-grounded evaluation with Vision-Language Model (VLM) reasoning and selectively invokes physically-grounded deterministic rule functions after interpreting the environmental context. 
     To train and evaluate DriveJudge, we curate a large-scale dataset of 33,577 challenging driving samples with human annotations on whether the driving behavior is reasonable in the given scenario.
     With this dataset, we address the underexplored problem of driving metric evaluation, and introduce two human-aligned benchmark tasks: Driving Quality Classification and Trajectory Preference Selection.
     DriveJudge outperforms EPDMS for driving quality classification by 21.23 AUC, and the recent VLM-based DriveCritic for trajectory preference selection by 6.5\%, setting a new standard for interpretable and precise driving evaluation.
\end{abstract}

\section{Introduction}
Autonomous driving has increasingly shifted towards end-to-end policy learning~\cite{li2024hydra,li2025hydranext,yao2025drivesuprim,li2025ztrs,li2025generalized,yao2026had,liao2024diffusiondrive,zou2025diffusiondrivev2,hu2023planning,chen2024vadv2}, which already achieves strong performance on the abundance of nominal driving scenarios~\cite{caesar2020nuscenes,dauner2024navsim,nvidia_physicalai_autonomous_vehicles_2025}. The next critical challenge for improving policies lies in the long tail, which entails non-nominal rare events and complex interactions~\cite{li2022coda,xu2025wod,alhaija2025cosmos,ren2025cosmos,chen2025eccv}.
Problematically, policy evaluations themselves are underdeveloped for long-tail scenarios: Existing metrics often fail to faithfully capture driving quality in long-tail settings, where contextual understanding of maneuvers is essential~\cite{abouelazm2026reviewrewardfunctionsreinforcement,chen2024criteria,hao2026styledrive,song2025drivecritic}.
Meaningful progress in autonomous driving ultimately depends on reliable driving quality assessment, which is hindered by a current lack of long-tail-ready driving metrics.

Existing driving metrics can be broadly categorized into imitation-based metrics~\cite{xu2025wod} and rule-based metrics~\cite{dauner2024navsim,dosovitskiy2017carla,li2022metadrive,jia2024bench2drive,alpasim_2025}. \emph{Imitation-based metrics}, such as Average/Final Displacement Error (ADE/FDE), evaluate trajectories by measuring similarity to expert demonstrations. 
Assuming a fixed correct solution makes them simple and scalable, but ignores that expert trajectories may be suboptimal or unavailable, and overlooks the inherently multi-modal nature of driving.

\emph{Rule-based metrics}, such as Extended Predictive Driver Model Score (EPDMS)~\cite{dauner2024navsim,cao2025pseudo}, 
assess driving quality through explicitly computed rules.
Because they provide transparent, reproducible, and fine-grained evaluation across dimensions like collision avoidance and lane compliance, rule-based metrics are the dominant paradigm for driving evaluation in both open-loop and closed-loop settings~\cite{dauner2024navsim,dosovitskiy2017carla,li2022metadrive,jia2024bench2drive,alpasim_2025}.

However, predefined rules are inherently context-agnostic. In long-tail scenarios, driving quality is often highly situational: Nudging around obstacles to maintain progress may require  temporary relaxation of lane-keeping constraints. Without contextual understanding, rule-based metrics can over-penalize such reasonable behaviors and produce misaligned evaluations. 

To address the missing context awareness of existing driving metrics, we introduce \textbf{\method{}}, a VLM-guided and rule-grounded evaluation agent for autonomous driving (Figure~\ref{fig:teaser}). \method{} formulates driving evaluation as an agentic decision-making process, where a VLM interprets the scene context and selectively invokes deterministic rule functions, such as collision checks and lane deviation analysis. This design bridges semantic reasoning and rule-based evaluation: the VLM provides contextual understanding and tool selection, while rule functions preserve the spatial precision and physical grounding required for reliable safety assessment.

\method{} differs fundamentally from existing VLM-based critic methods across autonomous driving, robotics, and other decision-making domains~\cite{song2025drivecritic,hao2026styledrive,lee2026roboreward,liang2026robometer,wu2026large,yi2026critic,guan2024task,xie2023text2reward,krack2026rewarding,xiong2026phycritic,langerman2025explaining,zhang2025critic,aghzal2025evaluating,zhang2026vq}, which mainly rely on either preference-based comparison or direct score prediction. Both paradigms depend on VLM-only judgments, making it difficult to explicitly decompose and interpret multiple driving aspects (e.g., safety, comfort, and progress), while also remaining unreliable for fine-grained safety assessment due to limited spatial precision~\cite{dauner2024navsim,wang2026vggdrive,luo2026last,li2025spacedrive,schmalfuss2025parc,huang20253d}. In contrast, \method{} introduces \textit{VLM-guided tool invocation}, combining contextual VLM reasoning with rule-grounded execution for context-aware, interpretable, and physically grounded driving evaluation.

To compare \method{} against baseline metrics, we argue that evaluating a driving metric itself is an underexplored problem, and propose two downstream benchmarks for metric evaluation (Sec~\ref{sec:metric_evaluation}): Driving Quality Classification following ~\cite{guan2024task} and Trajectory Preference Selection following~\cite{song2025drivecritic,langerman2025explaining,aghzal2025evaluating}. 
To create these benchmarks, and train and validate \method{}, we curate a large-scale dataset of 33,577 long-tail driving samples (Sec~\ref{sec:dataset}), where reasonable driving may require temporarily relaxing certain rule constraints. Our data focus on cases where human driving itself violates or underperforms on specific rule-based metrics, paired with human annotations on whether the behavior remains ultimately reasonable. To further improve \method{} (Sec~\ref{sec:SFT}), we develop a post-training pipeline combining supervised fine-tuning (SFT) and reinforcement learning (RL) to strengthen contextual reasoning and rule invocation. Experiments show that \method{} significantly outperforms prior state-of-the-art driving metrics, improving driving quality classification by \textbf{21.23 AUC over EPDMS}~\cite{dauner2024navsim} and human preference alignment by \textbf{6.5 accuracy over DriveCritic}~\cite{song2025drivecritic}, despite DriveCritic being a true preference model.
In summary, our contributions are as follows:

\begin{itemize}
\item We propose \method{}, a context-aware driving evaluation agent combining VLM-based reasoning with rule-grounded evaluation, enabling context-sensitive assessment of driving.
\item We curate and release a large-scale driving dataset with human annotations on complex driving maneuvers, labeling whether the observed behavior is ultimately reasonable. This dataset enables the study of driving evaluation and may benefit broader downstream applications.
\item We formulate the underexplored problem of \emph{evaluating driving metrics} and introduce two benchmark tasks---Driving Quality Classification and Trajectory Preference Selection---to validate metrics, and show \method{} significantly outperforms existing driving metrics.
\item We benchmark state-of-the-art driving policies under \method{}, revealing stronger driving capability than conventional metrics suggest and signs of benchmark saturation, highlighting the need for future research beyond incremental rule-based metric optimization.
\end{itemize}

\section{Related Work}
\label{sec:related}

\paragraph{Driving Evaluation Metrics.}
Driving evaluation has received far less attention than model and data development, and existing methods mainly fall into two categories: imitation-based and rule-based metrics. Imitation-based metrics, such as ADE/FDE and Waymo’s RFS~\cite{xu2025wod}, measure similarity to expert trajectories, assuming a fixed correct solution and relying heavily on demonstration quality, which limits reliability under multi-modal or imperfect human behaviors. As a result, the field has shifted toward rule-based metrics. In closed-loop settings~\cite{dosovitskiy2017carla,alpasim_2025,li2022metadrive,jia2024bench2drive}, these typically take the form of infraction scores. More recently, open-loop benchmarks such as NAVSIM and NAVSIM v2 have established PDMS and EPDMS as widely adopted interpretable and physically grounded driving metrics~\cite{dauner2024navsim,cao2025pseudo}.

However, rule-based metrics remain context-agnostic: in long-tail scenarios, reasonable driving may require relaxing certain rules (e.g., nudging around obstacles), leading to over-penalization. \method{} addresses this limitation by introducing a context-aware evaluation framework that uses VLMs to reason about rule applicability.
\paragraph{Evaluating an Evaluation Metric.}
To validate \method{}, a key challenge is to rigorously evaluate metric quality. We survey strategies across autonomous driving and other fields, including robotics~\cite{lee2026roboreward,liang2026robometer,guan2024task,wu2026large}, reconstruction~\cite{langerman2025explaining}, and more~\cite{aghzal2025evaluating}, and identify four common paradigms.
(1) \textit{Action Preference Selection}~\cite{song2025drivecritic,xiong2026phycritic,liang2026robometer,langerman2025explaining,aghzal2025evaluating}: a metric should assign higher scores to human-preferred actions, typically evaluated in pairwise settings.  
(2) \textit{Deployment Performance}~\cite{lee2026roboreward}: using the metric as a reward or model selection criterion should improve real-world performance, though this is often costly and impractical in autonomous driving.  
(3) \textit{Closed-loop Alignment}~\cite{dauner2024navsim,cao2025pseudo}: stronger open-loop metrics should correlate with closed-loop performance, though closed-loop metrics themselves may be imperfect.  
(4) \textit{Action Quality Classification}~\cite{guan2024task}: the metric should distinguish reasonable actions from failures.

Given the limitations of (2)–(3), we adopt (1) Action Preference Selection and (4) Action Quality Classification, which provide direct, scalable, and reliable measures of quality of driving metric.

\paragraph{VLM-based Critic and Judge Models.}
Recent work explores Vision-Language Models (VLMs) as critics for policy evaluation, which mainly fall into two paradigms. The first is \textit{preference-based models}~\cite{song2025drivecritic,xiong2026phycritic,langerman2025explaining,liang2026robometer,aghzal2025evaluating}, where VLMs select the better action among candidates. While effective for relative ranking, they do not produce explicit quality scores, limiting usage as standalone metrics or reward functions over larger candidate sets. The second is \textit{direct scoring}~\cite{lee2026roboreward,zhang2026vq,wu2026large,zhang2025critic}, where VLMs directly predict scalar scores or textual feedback for evaluating actions.

Despite their promise, both paradigms share two key limitations. First, autonomous driving is inherently multi-objective~\cite{dauner2024navsim}, requiring joint assessment of safety, progress, comfort, and rule compliance, making VLM-only judgments ambiguous and less interpretable. Second, VLMs remain limited in spatial precision~\cite{wang2026vggdrive,luo2026last,li2025spacedrive,huang20253d}, making direct fine-grained safety assessment unreliable.

In contrast, \method{} introduces a third paradigm: \textit{VLM-guided tool invocation}, where VLMs first understand the context then perform contextual reasoning to select evaluation tools, while deterministic rule functions provide spatially precise and physically grounded assessment.

\paragraph{VLM Post-Finetuning}
Adapting Vision-Language Models (VLMs) to specialized domains typically relies on a two-stage post-training pipeline consisting of supervised fine-tuning (SFT)~\cite{ouyang2022training,wei2021finetuned,chung2024scaling,liu2023visual} followed by reinforcement learning (RL)~\cite{schulman2017proximal,deepseekai2025deepseekv3,liu2026gdpo,liu2025dapo}. SFT aligns the model with task-specific formats, reasoning patterns, and domain knowledge, while RL further optimizes objectives that are difficult to supervise directly, such as preference alignment and decision-making quality. In this work, we adopt this paradigm to improve \method{} for context-aware rule invocation and human-aligned driving evaluation. Specifically, we employ GRPO~\cite{deepseekai2025deepseekv3} during the RL stage due to its strong sample efficiency and effectiveness for preference-based optimization over multiple candidate rollouts.


\section{How to Evaluate a Driving Metric?}
\label{sec:metric_evaluation}

Following evaluation paradigms in related fields (Sec.~\ref{sec:related}), a good metric should align with human judgment. We therefore propose to evaluate driving metric via two human-aligned downstream tasks.

\subsection{Driving Quality Classification}
First, following ~\cite{guan2024task}, a good driving metric should distinguish reasonable from unreasonable driving, assigning high scores to human-approved behaviors and low scores to failures. Similar strategies have been used to evaluate VLM-based critics in robotics~\cite{guan2024task}. Given a binary label $y \in \{0,1\}$ and a metric score $\mathcal{S} \in [0,1]$, we classify a behavior as reasonable if $\mathcal{S} > \tau$:
\begin{align}
\label{eq:tau}
\hat{y}_i =
\begin{cases}
1, & \mathcal{S}_i > \tau \\
0, & \text{otherwise.}
\end{cases}
\end{align}
The \textbf{D}riving \textbf{Q}uality \textbf{C}lassification accuracy is:
\begin{align}
\label{eq:acc_dqc}
\mathrm{ACC}_{\mathrm{DQC}} = \frac{1}{N}\sum_{i=1}^{N}\mathds{1}[\hat{y}_i = y_i].
\end{align}
Since accuracy depends on $\tau$, we report $\mathrm{ACC}_{\mathrm{DQC}}$ across multiple thresholds. 
To provide threshold-independent evaluation, we also report AUC and average precision: $\mathrm{AUC}(S, \mathbf{y}), \quad \mathrm{AP}^{+}(S, \mathbf{y}).$ To emphasize safety-critical failure detection, we additionally report: $\mathrm{AP}^{-}(1-S, 1-\mathbf{y})$, which measures average precision on detecting failures. 
We evaluate on a validation split of 1083 samples from the \method{} dataset(Sec~\ref{sec:dataset}), with human annotations as ground-truth labels.

\subsection{Trajectory Preference Selection}
Second, a good driving metric should align with human preference when comparing candidate actions. Given two trajectories $\mathrm{traj}^A$ and $\mathrm{traj}^B$ with human preference label $y \in \{A, B\}$, the metric should assign a higher score to the preferred trajectory. Similar protocols have been used in robotics, 3D reconstruction, and physical decision-making~\cite{liang2026robometer,langerman2025explaining,xiong2026phycritic}. The evaluation accuracy is defined as:
\begin{align}
\label{eq:acc_tps}
\mathrm{ACC}_{\mathrm{TPS}} = \frac{1}{N} \sum_i^N \mathds{1}[\arg\max(\mathcal{S}^A_i, \mathcal{S}^B_i) = y_i].
\end{align}
We evaluate on the DriveCritic~\cite{song2025drivecritic} validation set for NAVSIM~\cite{dauner2024navsim} with pairwise trajectory preferences annotated by GPT-4~\cite{openai2024gpt4} and verified by humans. Ambiguous cases without clear preference are further removed by us, resulting in 862 samples (see details in Appendix~\ref{subsec:navsim_filtering}).

\section{\method{}}
\label{sec:drivejudge}

We propose \method{}, a context-aware driving evaluation agent that leverages VLM reasoning to interpret scene context and selectively invoke rule-grounded evaluation functions.

\begin{figure}[t]
    \centering
    \includegraphics[width=.85\linewidth]{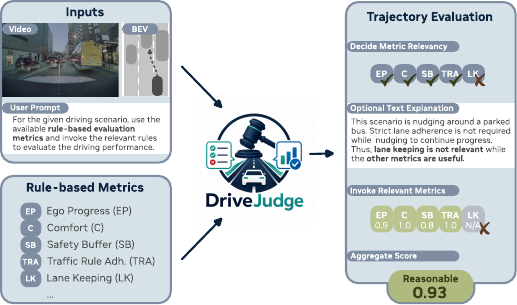}
    \caption{\textbf{\method{} Framework.} \method{} leverages a VLM to determine metric relevance under scene context, selectively invokes the relevant rules, and aggregates their outputs into a context-aware driving quality score.}
    \label{fig:data_method_METHOD}
\end{figure}

\begin{figure}[t]
    \centering
    \includegraphics[width=.85\linewidth]{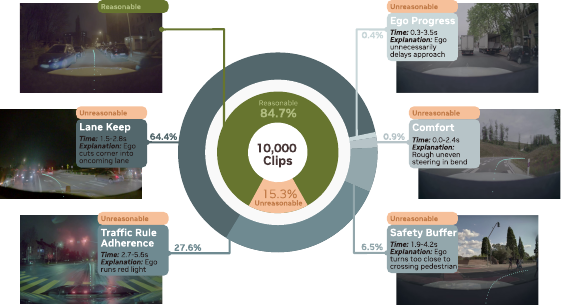}
    \caption{\textbf{\method{} Dataset.} The \method{} dataset contains diverse challenging driving scenarios, including reasonable behaviors that may deviate from conventional driving rules, as well as fine-grained driving failure cases.}
    \label{fig:data_method_DATA}
\end{figure}

\subsection{DriveJudge Score via Context-Aware Rule Invocation}

\method{} formulates driving evaluation as an agentic process, where a VLM interprets scene context and selectively invokes rule-based functions for physically grounded and spatially precise assessment (Fig.~\ref{fig:data_method_METHOD}).
Let $\mathcal{M}=\{f^1,f^2,\dots,f^M\}$ denote a set of rule-based evaluation functions, where each rule evaluates a driving quality dimension (e.g., safety, progress, etc.) to produce a compliance score:
\begin{align}
s^m = f^m(\cdot), \quad s^m \in [0,1].
\end{align}
Conventional rule-based driving metrics~\cite{dauner2024navsim,cao2025pseudo} aggregate all rule scores $s^m$ into a weighted composite score that implicitly assumes equal rule relevance in every driving scenario, where $w^m$ is the weight:
\begin{align}
\mathcal{S}^{\text{conv}}_i = \frac{\sum_{m \in \mathcal{M}} w^m_i s^m_i}{\sum_m w^m_i}.
\end{align}
We argue that rule relevance is inherently context-dependent. For instance, in nudging scenarios, temporarily relaxing lane-keeping constraints may be necessary for safe obstacle avoidance and maintaining progress. To address this, \method{} dynamically determines which rules should contribute to evaluation based on scene context.
Specifically, \method{} predicts a gating vector $\hat{\mathbf g}_i$, where each gating variable $\hat g_i^m \in [0,1]$ indicates the relevance of rule $f^m$ for driving state $X_i$:
\begin{align}
\label{eqn:drivejudge}
\hat{\mathbf g}_i = \Theta_{\mathrm{DJ}}(X_i, \text{prompt}),
\end{align}
where $\Theta_{\mathrm{DJ}}$ denotes the \method{} agent, and we show the prompt details in Appendix \ref{subsec:prompt}.
Given a predicted trajectory $\hat t_i$, the selected rules are evaluated as:
\begin{align}
s_i^m = f^m(X_i, \hat t_i).
\end{align}
The final \method{} driving score is computed by:
\begin{align}
\label{eqn:drivejudge_score}
\mathcal{S}^{\mathrm{DJ}}_i=
\frac{\sum_{m \in \mathcal{M}}
\hat g_i^m w^m s_i^m}
{\sum_{m \in \mathcal{M}}
\hat g_i^m w^m}
\end{align}
\method{} is agnostic to the choice of rule set, as long as it captures the core aspects of driving. In this work, we instantiate $\mathcal{M}$ using EPDMS~\cite{dauner2024navsim}, with rule functions including safety (NC, TTC), lane compliance (LK, DAC, DDC), progress (EP), and comfort.

For input, each driving state $X_i$ consists of camera observations and perception labels (e.g., object boxes and lane maps), where the latter are rendered as bird eye's view (BEV) visualizations and provided to the VLM together with camera images as multi-modal input. 

To train and validate \method{}, we introduce a curated driving evaluation dataset next.
\subsection{DriveJudge Dataset}
\label{sec:dataset}

Long-tail and complex driving scenarios are critical for evaluating autonomous systems and studying driving metrics, but yet remain underrepresented in existing benchmarks~\cite{nvidia_physicalai_autonomous_vehicles_2025,dauner2024navsim,cao2025pseudo}, which largely focus on nominal driving. Therefore, we curate a dataset from the publicly available PhysicalAI-Autonomous-Vehicles dataset~\cite{nvidia_physicalai_autonomous_vehicles_2025}, which contains over $300{,}000$ driving clips of $20$ seconds long.

To identify informative samples for metric learning, we compute rule-based metric scores on human trajectories using the EPDM~\cite{dauner2024navsim} metric suite and mine challenging segments where human driving receives low rule-based scores, capturing ambiguous and context-dependent behaviors. Human annotators are then asked a binary question: \textit{Does the driving behavior follow reasonable driving practices under the current scene context?} They also provide brief textual justifications for their decisions.

In total, we label $10,732$ segments and extract $33,577$ standardized $2$s windows, with $15.3\%$ labeled as driving failures (Fig.~\ref{fig:data_method_DATA}). Those driving failures violate at least one of the five classes: lane keep, traffic rule adherence, ego progress, comfort or safety buffer. The dataset is split into training and a representative test set of $1,083$ samples with a higher failure ratio ($23.3\%$). This dataset serves both as supervision for training \method{} and as a benchmark for evaluating driving metrics.
\subsection{\method{} Finetuning}
We further improve \method{} using a two-stage post-training pipeline of SFT followed by RL.
\label{sec:SFT}
\paragraph{Supervised Fine-Tuning (SFT)}
The SFT stage directly supervises \method{} on rule invocation decisions, teaching \method{} whether each evaluation rule should be invoked for a given driving clip. This provides an efficient way to improve the context-aware tool-calling ability of \method{}.

To construct ground-truth rule invocation labels, we leverage the human annotations in the \method{} dataset. Each clip is annotated with a binary label $y_i \in \{0,1\}$, where $y_i=1$ indicates reasonable driving and $y_i=0$ indicates unreasonable driving.
Our key intuition is that if a human trajectory is judged as reasonable but receives a low score under a rule function, that rule is likely irrelevant in the given context rather than indicating an actual failure. For example, in emergency stopping scenarios, metrics such as Ego Progress or Comfort may score poorly while the behavior remains reasonable.
Formally, for each sample $i$, we first compute rule-based scores $s_i^m$ on the human trajectory $t_i$, from which we define the ground-truth rule invocation label $g_i^m$:
\begin{align}
s_i^m &= f^m(P_i, t_i), \quad m \in \mathcal{M}, \\
g_i^m &=
\begin{cases}
0, & y_i = 1 \text{ and } s_i^m < \tau^m \\
1, & \text{otherwise}.
\end{cases}
\end{align}
Here, $\tau^m$ is the violation threshold for rule $m$. We select these thresholds to maximize alignment between rule-based outcomes and human judgments on the \method{} dataset (details in Sec.~\ref{subsec:thresholds}).
We then optimize \method{} with a supervised cross-entropy loss:
\begin{align}
\mathcal{L}_{\mathrm{SFT}} =
\sum_{i=1}^{N} \sum_{m \in \mathcal{M}}
\mathrm{CE}(\hat{g}_i^m, g_i^m).
\end{align}

\paragraph{Reinforcement Learning}
As discussed in Sec.~\ref{sec:metric_evaluation}, a strong driving metric should perform well on downstream tasks including driving quality classification and trajectory preference selection. Since SFT labels are derived from human driving quality annotations, we expect SFT alone to already provide strong performance on quality classification. We therefore introduce a reinforcement learning (RL) stage to further optimize \method{} for improving trajectory preference selection.

For each preference sample $i$, two candidate trajectories $\mathrm{traj}_i^A$ and $\mathrm{traj}_i^B$ are paired with a human preference label $y_i \in \{A,B\}$. \method{} evaluates both trajectories and produces driving scores $\mathcal{S}_i^A$ and $\mathcal{S}_i^B$. The reward is defined by whether the metric ranking matches human preference:
\begin{align}
\label{eqn:reward}
r_i =
\begin{cases}
1, & \arg\max(\mathcal{S}_i^A, \mathcal{S}_i^B) = y_i, \\
0, & \text{otherwise}.
\end{cases}
\end{align}

To optimize this objective, we adopt Group Relative Policy Optimization (GRPO)~\cite{deepseekai2025deepseekv3}, which optimizes relative advantages over sampled rule-invocation rollouts. With $|\mathcal{M}|=7$ rules, the action space contains $128$ possible invocation combinations; we use a rollout size of $8$ for exploration.

\section{Experiments}
We evaluate \method{} under the validation protocol in Sec.~\ref{sec:metric_evaluation} against existing driving metrics on two downstream tasks: Driving Quality Classification and Trajectory Preference Selection. We report quantitative and qualitative results, ablations on VLM-guided tool invocation, reinforcement learning, and scaling behaviors under varying SFT and RL data sizes, and further benchmark state-of-the-art policy models under \method{}.
For implementation, we finetune Qwen-3~\cite{yang2025qwen3} (2B and 8B) using the VERL framework~\cite{sheng2024hybridflow}. SFT trains models for 9,800 iterations with batch size 32 and learning rate $10^{-5}$. RL optimizes models with GRPO for 340 iterations with batch size 32, rollout size 8, learning rate $10^{-6}$, and KL coefficient $10^{-2}$. All experiments ran on 8 NVIDIA A100 GPUs.

\begin{table}[t!]
\centering
\caption{\textbf{Driving Quality Classification Performance.}
We report $\mathrm{ACC}_{\mathrm{DQC}}$ (Eq.~\ref{eq:acc_dqc}) under multiple thresholds $\tau$ (Eq.~\ref{eq:tau}), along with threshold-independent ranking metrics including AUC and Average Precision for reasonable ($\mathrm{AP}^{+}$) and unreasonable ($\mathrm{AP}^{-}$) driving (Sec.~\ref{sec:metric_evaluation}). Across all settings, \method{} consistently outperforms strong baselines by a significant margin. Results are run twice.}
\vspace{3pt}
\label{tab:quality_classification}
\begin{tabular}{l c | ccc c c c}
\toprule
\multirow{2}{*}{Method} & \multirow{2}{*}{Finetuning} & \multicolumn{3}{c}{Accuracy ($\mathrm{ACC}_{\mathrm{DQC}}$)} & \multirow{2}{*}{AUC} & \multirow{2}{*}{\makecell{$\mathrm{AP}^-$} } & \multirow{2}{*}{\makecell{$\mathrm{AP}^+$} } \\
\cmidrule(lr){3-5}
 & & $\tau=0.7$ & $\tau=0.8$ & $\tau=0.9$ & & \\
\midrule
Random & \xmark      & 50.00 & 50.00 & 50.00 & 50.00 & 23.3 & 76.7 \\
ADE/FDE & \xmark      & n/a & n/a & n/a & n/a & n/a & n/a \\
EPDMS~\cite{cao2025pseudo} & \xmark      & 66.94 & 54.76 & 43.95 & 57.67 & 26.34 & 81.64\\
Gemini3.1~\cite{comanici2025gemini} & \xmark  & 65.73 & 65.69 & 65.26 & 56.61 & 26.29 & 79.62 \\
GPT-5.4~\cite{openai2024gpt4} & \xmark      & 76.71 & 76.59 & 72.48 & 48.59 & 23.65 & 76.89 \\
Supervised Classifier & \cmark  & 77.01 & 73.59 & 33.51 & 58.39 & 33.28 & 80.49 \\
\rowcolor{lgreen}
\modelName-2B & \cmark & \textbf{80.97} & 83.28 & 83.74 & 77.95 & 58.31 & 89.83 \\
\rowcolor{lgreen}
\modelName-8B & \cmark  & 80.89 & \textbf{83.47} & \textbf{84.03} & \textbf{78.90} & \textbf{59.04} & \textbf{90.77}\\
\bottomrule
\end{tabular}
\vspace{-3pt}
\end{table}

\begin{figure}[t!]
    \centering
    \begin{minipage}[t]{0.48\textwidth}
        \centering
        \subcaptionbox{Direct scoring misses the lane-keeping violation during the left turn due to weak spatial precision. In contrast, \method{} invokes rule-based evaluation and assigns a more appropriate driving score of 0.70. \label{fig:error_example_swingwide_leftturn}}{
            \includegraphics[width=\textwidth]{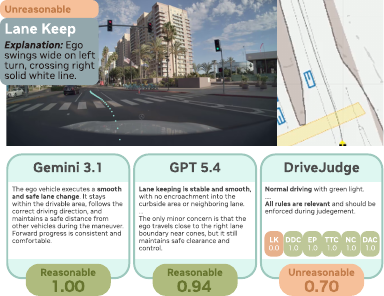}
        }
    \end{minipage}
    \hfill
    \begin{minipage}[t]{0.48\textwidth}
        \centering
        \subcaptionbox{Ego nudges into the opposite lane to avoid a construction zone. While EPDMS heavily penalizes this behavior, \method{} correctly recognizes the contextual justification and assigns a driving score of 1.00. \label{fig:error_example_nudge}}{
            \includegraphics[width=\textwidth]{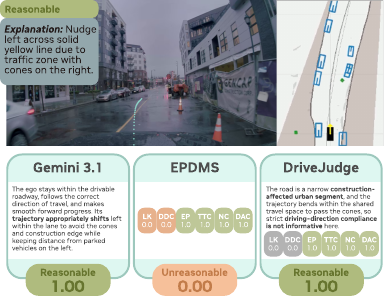}
        }
    \end{minipage}
    \caption{\textbf{Qualitative model comparisons} on two representative clips from the test set.}
    \label{fig:error_examples_combined}
\end{figure}

\begin{figure*}[t]
\centering

\begin{minipage}[t]{0.48\textwidth}
\centering
\captionof{table}{\textbf{Trajectory Preference Selection Performance.}
We report $\mathrm{ACC}_{\mathrm{TPS}}$ (Eq.~\ref{eq:acc_tps}) to compare \method{} against strong baselines, including EPDMS and DriveCritic, which is an explicit preference model. \method{} consistently outperforms all baselines by a significant margin in selecting trajectories aligned with human preference. Results are the mean of two repetitions.}
\vspace{3pt}
\label{tab:trajectory_selection}
\small
\begin{tabular}{l c | c}
\toprule
Method & Finetuning & Accuracy \\
\midrule
Random & \xmark & 50.00 \\
ADE/FDE & \xmark & 70.53 \\
EPDMS~\cite{cao2025pseudo} & \xmark & 56.50 \\
GPT-5~\cite{openai2024gpt4} & \xmark & 55.20 \\
Gemini3.1~\cite{comanici2025gemini} & \xmark & 55.45 \\
Supervised Classifier & \cmark & 64.80 \\
DriveCritic~\cite{song2025drivecritic} & \cmark & 76.33 \\
\rowcolor{lgreen}
\modelName-2B & \cmark & 80.63 \\
\rowcolor{lgreen}
\modelName-8B & \cmark & \textbf{82.83} \\
\bottomrule
\end{tabular}
\end{minipage}
\hfill
\begin{minipage}[t]{0.48\textwidth}
\centering
\captionof{figure}{\textbf{Qualitative comparison.} DriveCritic~\cite{song2025drivecritic} fails due to weak spatial grounding, while \method{} selects the correct trajectory.}
\includegraphics[width=.95\linewidth]{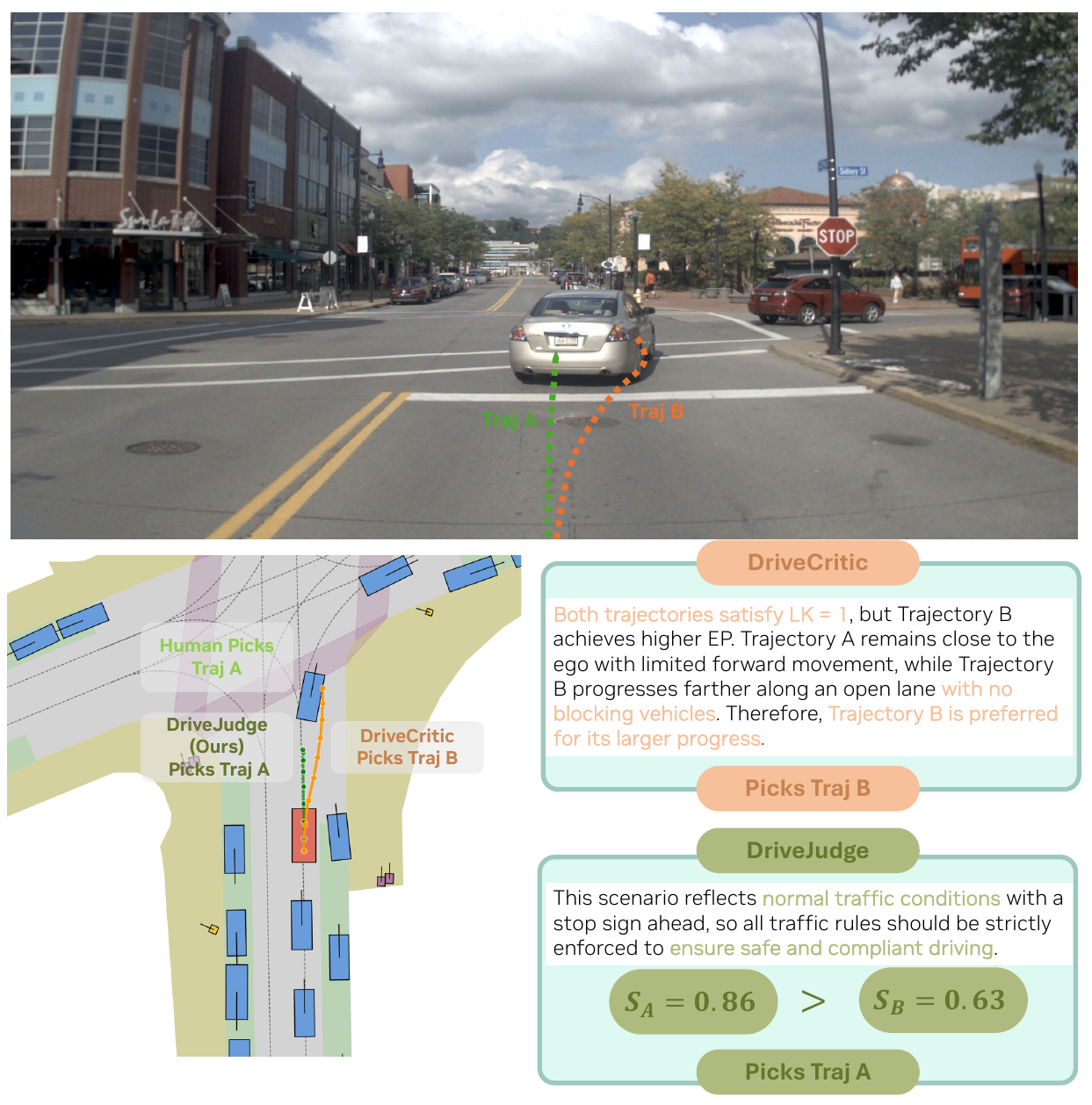}
\label{fig:navsim_fig}
\end{minipage}
\vspace{-10pt}
\end{figure*}

\paragraph{Baselines}
We compare \method{} against a diverse set of strong baselines across both downstream tasks. These include imitation-based metrics (ADE/FDE) and the widely adopted rule-based metric EPDMS~\cite{cao2025pseudo}. We further benchmark against general-purpose VLMs (GPT-5~\cite{openai2024gpt4}, Gemini 3.1~\cite{comanici2025gemini}) prompted for evaluation, as well as a trained non-VLM classifier (details in Sec~\ref{subsec:non_vlm_detail}) to isolate the benefit of VLM-based reasoning. For trajectory preference selection, we additionally compare against DriveCritic~\cite{song2025drivecritic}, a recent learned critic explicitly and directly designed for preference selection.




\subsection{Driving Quality Classification}


\paragraph{Quantitative Analysis}
In Table~\ref{tab:quality_classification}, we compare Driving Quality Classification performance of \method{} with all baselines. Imitation-based metrics (ADE/FDE) are not applicable, as they treat human trajectories as ground truth and therefore cannot identify failures or suboptimal human driving.

EPDMS~\cite{dauner2024navsim} performs poorly (57.67 AUC, 26.34 $\mathrm{AP}^{-}$) due to its rigid rule-based evaluation. Zero-shot VLMs (GPT~\cite{openai2024gpt4}, Gemini~\cite{comanici2025gemini}) show promise and achieve higher thresholded accuracy, but remain weaker in AUC. A supervised non-VLM classifier improves to 58.39 AUC and 33.28 $\mathrm{AP}^{-}$, validating the value of our curated dataset.
\method{} performs best across all metrics. Our 8B model reaches 78.90 AUC and 59.04 $\mathrm{AP}^{-}$, outperforming EPDMS by +21.23 AUC and +32.70 $\mathrm{AP}^{-}$. The strong gain in $\mathrm{AP}^{-}$ is particularly important for reliably detecting failure cases and highlights the effectiveness of contextual reasoning with rule-grounded evaluation.




\paragraph{Qualitative Analysis.}
Figure~\ref{fig:error_examples_combined} highlights representative failure modes of direct scoring VLMs and rule-based metrics. Direct scoring VLMs miss violations due to weak spatial fidelity, while rule-based metrics over-penalize contextually justified nudging. In contrast, \method{} combines contextual reasoning with rule-grounded evaluation, producing more accurate and robust assessments.

\subsection{Trajectory Preference Selection}
\paragraph{Quantitative Analysis}
In Table~\ref{tab:trajectory_selection}, we compare Trajectory Preference Selection performance across all baselines. EPDMS performs poorly on this task (56.50\%), reflecting the limitation of rigid rule-based metrics in aligning with human preferences. ADE/FDE achieves higher accuracy (70.53\%) by leveraging trajectory similarity, but remains limited by reliance on potentially imperfect human demonstrations. Zero-shot VLMs (GPT-5, Gemini) show some promise but still underperform, indicating that contextual reasoning alone is insufficient without physical grounding.

\method{} achieves the best performance, outperforming all baselines by a clear margin. Notably, it surpasses DriveCritic~\cite{song2025drivecritic} (82.83\% vs. 76.33\%), despite DriveCritic being designed only for preference-based selection. We attribute this gap to our design: while both use VLMs for contextual reasoning, \method{} further incorporates rule-based tool invocation to ensure spatial and physical fidelity, which is critical for reliable trajectory evaluation.

\paragraph{Qualitative Analysis}
Figure~\ref{fig:navsim_fig} shows a qualitative comparison between DriveCritic~\cite{song2025drivecritic} and \method{}. Due to weak spatial understanding, DriveCritic makes factual errors: it incorrectly claims there is no blocking vehicle ahead, treats moving over for progress as appropriate, and assumes both trajectories satisfy lane keeping. These errors lead it to prefer Trajectory B, contradicting the human label. In contrast, \method{} grounds evaluation through explicit rule invocation, correctly identifies the traffic condition, and selects the human-preferred Trajectory A.

\begin{wraptable}{r}{0.5\columnwidth}
\vspace{-8pt}
\centering
\small
\caption{
Ablation on VLM reasoning, tool invocation, and finetuning.
}
\vspace{-5pt}
\label{tab:ablation_main}
\setlength{\tabcolsep}{3.5pt}
\begin{tabular}{l|cc|c|c}
\toprule
Method & VLM & Tool & FT & AUC \\
\midrule
VLM-DirectScore      & \cmark & \xmark & \xmark & 50.06 \\
\method{}-ZeroShot   & \cmark & \cmark & \xmark & 56.13 \\
\midrule
Non-VLM              & \xmark & \xmark & \cmark & 58.39 \\
Non-VLM-Agent        & \xmark & \cmark & \cmark & 55.11 \\
VLM-DirectScore-Ft   & \cmark & \xmark & \cmark & 68.67 \\
\rowcolor{lgreen}
\method{}            & \cmark & \cmark & \cmark & \textbf{77.95} \\
\bottomrule
\end{tabular}
\vspace{-10pt}
\end{wraptable}

\subsection{Ablation Studies}
\begin{figure}[t!]
    \centering
    \begin{minipage}[t]{0.32\linewidth}
        \centering
        \includegraphics[width=\linewidth]{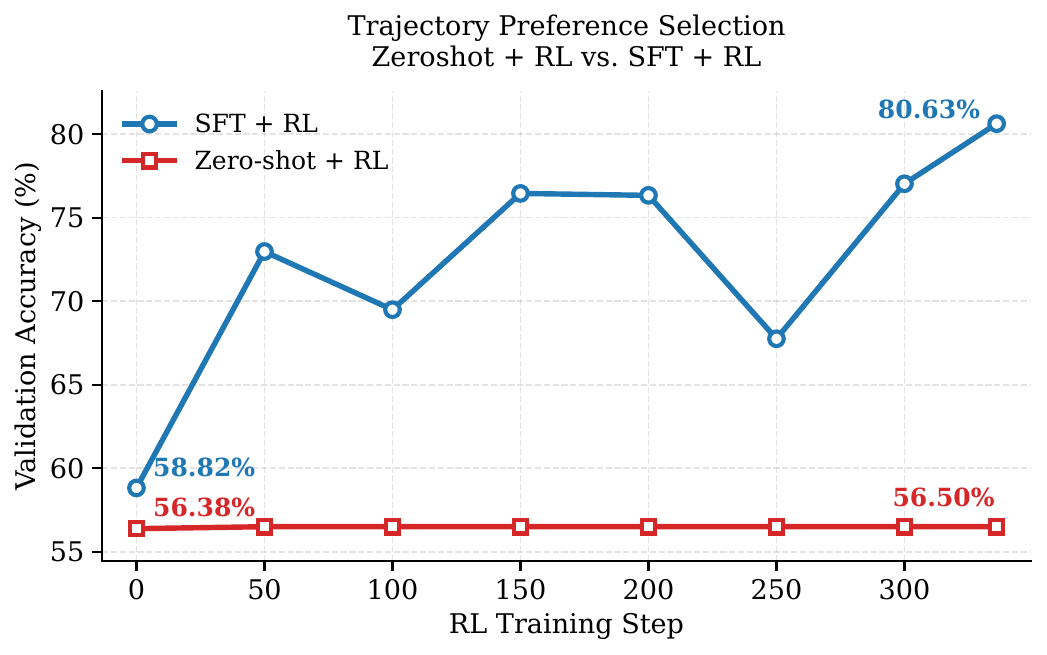}
        \caption{\textbf{Validation accuracy of Zero-shot+RL and SFT+RL} using \method{}-2B.}
        \label{fig:rl_training_dynamics}
    \end{minipage}
    \hfill
    \begin{minipage}[t]{0.65\linewidth}
        \centering
        \includegraphics[width=0.49\linewidth]{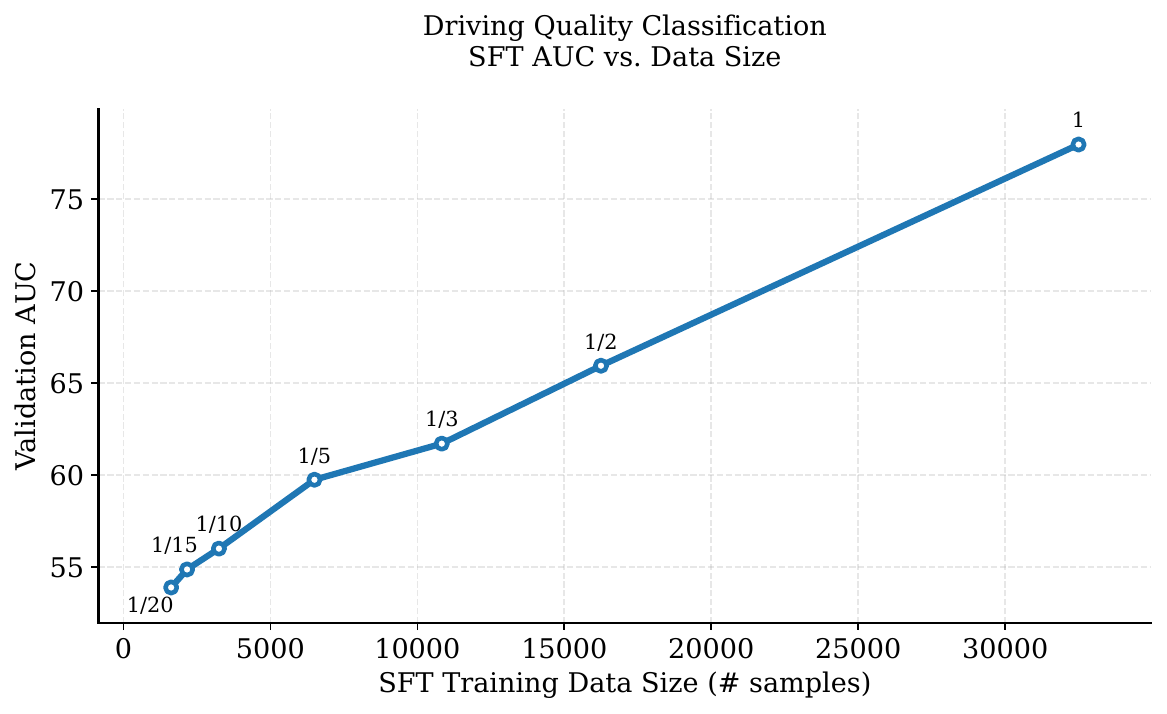}
        \includegraphics[width=0.49\linewidth]{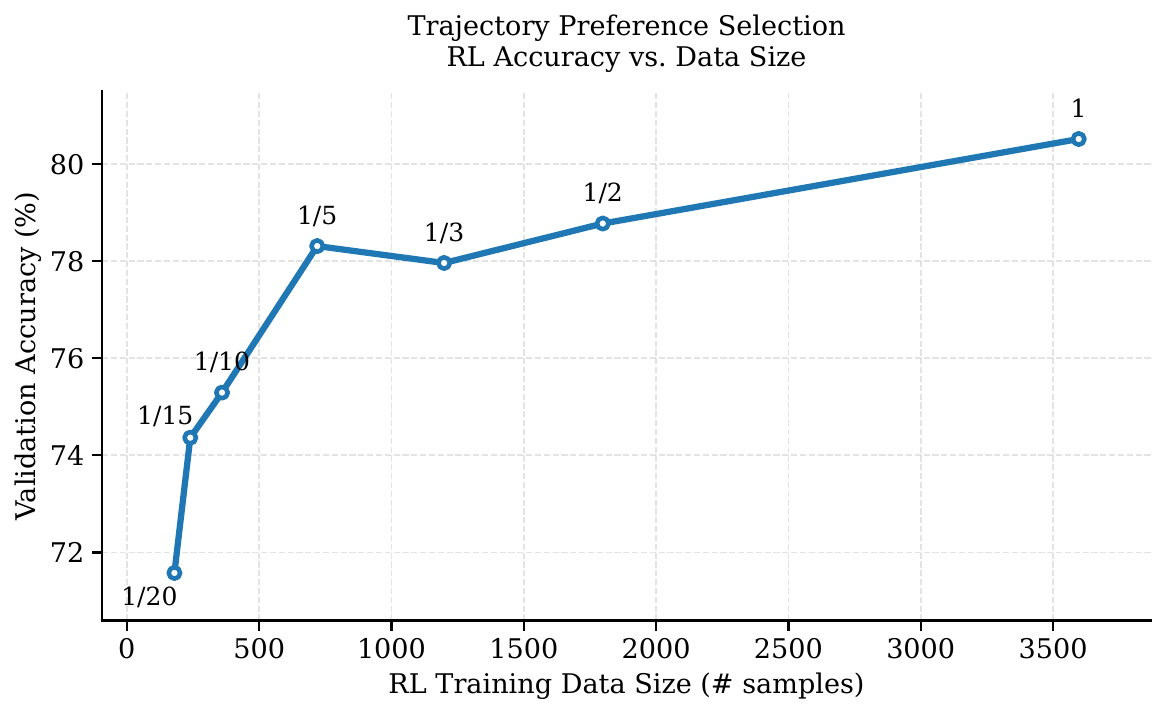}
        \caption{\textbf{Scaling behavior under varying training data sizes.} Left: Supervised Finetuning (SFT). Right: Reinforcement Learning (RL). All experiments use \method{}-2B.}
        \label{fig:data_size_ablation}
    \end{minipage}
\vspace{-5pt}
\end{figure}

\paragraph{Effectiveness of VLM, Tool Invocation, and Post-Finetuning.}
Table~\ref{tab:ablation_main} shows the contribution of three components: VLM reasoning, tool invocation, and post-finetuning. Tool invocation improves zero-shot performance (50.06 → 56.13 AUC), highlighting the need for explicit rule-based evaluation of spatially precise properties. Tasks such as collision detection and lane compliance require accurate geometric reasoning, which VLMs alone cannot reliably capture. Post-finetuning further amplifies this gap: \method{} improves to 77.95 AUC, while direct scoring reaches only 68.67 AUC. Although finetuning enhances contextual reasoning, it cannot overcome the inherent spatial fidelity limitations of VLMs without grounded tool support. Finally, finetuned non-VLM variants underperform both VLM-based baselines, emphasizing the importance of VLM commonsense and contextual understanding. Overall, combining VLM reasoning with rule-grounded tool invocation is key to strong performance.
\paragraph{Reinforcement Learning for Trajectory Preference Selection.}
In Fig.~\ref{fig:rl_training_dynamics}, we compare validation accuracy during RL optimization, starting from either an SFT-initialized model or a zero-shot model under identical settings. RL on top of SFT leads to large gains in validation accuracy (58.82\% → 80.63\%). In contrast, starting from zero-shot yields nearly no change in accuracy (56.38\% → 56.50\%), indicating minimal learning. This suggests RL alone is insufficient for effective exploration from a zero-shot initialization. SFT provides a critical warm start by grounding the model in task structure and reasoning, enabling RL to further refine preference-aligned decision making.
\paragraph{Scaling Behavior Under Training Data Size.}
In Fig.~\ref{fig:data_size_ablation}, we study how SFT and RL scale with training data. All runs use identical training iterations and batch sizes, and RL is initialized from the same fully SFT-trained model to isolate the effect of data scale. SFT shows steady, near-linear improvement with increasing data, indicating strong scalability with more annotated supervision. In contrast, RL achieves large gains at small data sizes (e.g., $1/20 \rightarrow 1/15$), followed by diminishing returns as performance saturates. This aligns with prior observations: SFT benefits from data scale~\cite{kaplan2020scaling,chung2024scaling}, while RL is more sensitive to data quality~\cite{ouyang2022training,ziegler2019fine}. Notably, even limited high-quality preference data yields substantial improvements, highlighting the importance of data quality in RL-based post-training for driving evaluation.

\begin{wraptable}{r}{0.5\linewidth}
\vspace{-10pt}
\centering
\small
\caption{
Comparison of state-of-the-art planning models on \texttt{navtest}~\cite{dauner2024navsim} under \method{}.
}
\vspace{-5pt}
\label{tab:planning_model_eval}
\begin{tabular}{lc>{\columncolor{lgreen}}c}
\toprule
Method & EPDMS $\uparrow$ & \method{} $\uparrow$ \\
\midrule
Hydra-MDP-8192~\cite{li2024hydra}  & 0.825 & 0.894 \\
Hydra-MDP-16384~\cite{li2024hydra} & 0.848 & 0.912 \\
\midrule
DriveSuprim-R34~\cite{yao2025drivesuprim}   & 0.818 & 0.882 \\
DriveSuprim-V2-99~\cite{yao2025drivesuprim} & 0.847 & 0.906 \\
DriveSuprim-ViT-L~\cite{yao2025drivesuprim} & 0.857 & 0.917 \\
\midrule
ZTRS-R34~\cite{li2025ztrs}   & 0.836 & 0.900 \\
ZTRS-V2-99~\cite{li2025ztrs} & 0.854 & 0.911 \\
ZTRS-ViT-L~\cite{li2025ztrs} & 0.865 & 0.924 \\
\bottomrule
\end{tabular}
\vspace{-10pt}
\end{wraptable}

\subsection{Benchmarking State-of-the-Art Policy Models}

To better understand modern autonomous driving policies under \method{} evaluation, we benchmark state-of-the-art planning models~\cite{li2024hydra,yao2025drivesuprim,li2025ztrs}. Table~\ref{tab:planning_model_eval} reports both EPDMS and DriveJudge scores. DriveJudge consistently assigns higher scores across all models than EPDMS, suggesting that rule-based metrics may undervalue contextually reasonable behaviors and underestimate the capability of modern driving policies. Meanwhile, the relatively small performance gaps among state-of-the-art models under DriveJudge indicate that benchmarks such as \texttt{navtest}~\cite{dauner2024navsim} may be approaching saturation.


\section{Conclusion}
\label{sec:conclusion}
\paragraph{Limitations.} A main limitation of this work is that we validate \method{} only in open-loop settings and do not conduct closed-loop experiments. Our current study focuses on open-loop metric quality. Also, an important future direction is to leverage \method{} as a reward function for policy optimization, where its context-aware and physically grounded metrics could provide richer supervision for training autonomous driving policies.

\paragraph{Conclusion.} We introduced \method{}, a context-aware driving evaluation agent that combines VLM reasoning with rule-grounded evaluation for autonomous driving. By selectively invoking evaluation rules based on scene context, \method{} enables interpretable, physically grounded, and human-aligned assessment of driving quality, particularly in long-tail scenarios where conventional metrics often fail. We further curated a large-scale benchmark for evaluating driving metrics and showed that \method{} consistently outperforms existing metrics on both driving quality classification and trajectory preference alignment. We hope this work advances the study of driving evaluation and provides a stronger foundation for future autonomous driving research.

{
    \small
    \bibliographystyle{plain}
    \bibliography{main}
}
\newpage
\appendix

\section{Model Details}
\subsection{Prompts}
\label{subsec:prompt}

\lstset{
    basicstyle=\small\ttfamily,
    breaklines=true,
    frame=single,
    captionpos=b,
    backgroundcolor=\color{gray!10}
}

\begin{lstlisting}[caption={DriveJudge Prompt}]
System: You are an expert AI driving instructor and analyst.
Your task is to analyze which criteria are useful for evaluating driving behavior in the given video context.
Focus on the following evaluation criteria:
    - Lane Keeping (LK): A lower score indicates deviation from the lane centerline, while a higher score means the vehicle stays well-centered within its lane. Ego vehicle should drive as close to lane centerlines as possible. However, LK can be "too strict" during justified lane changes, courteous nudges, overtakes, construction work zones, or wide lanes; question LK if the camera view contradicts it.
    - Driving Direction Compliance (DDC): A lower score means the vehicle is not following the intended lane direction (e.g., driving in the opposite direction). DDC can be "too strict" during courteous nudging or construction work zones; question DDC if the camera view contradicts it.
    - Drivable Area Compliance (DAC): A lower score indicates the vehicle is moving outside the designated drivable area. DAC can be "too strict" during courteous nudging or work zones; question DAC if the camera view contradicts it.
    - Time to Collision (TTC): A lower score suggests the trajectory may lead to an imminent collision with other agents or pedestrians. This should always be critical as there are agents, pedestrains, or obstacles visible.
    - No Collision (NC): A lower score indicates the trajectory results in a collision with other agents or pedestrians. This should always be critical as there are agents, pedestrains, or obstacles visible.
    - Ego Progress (EP): A lower score reflects poor navigation progress toward the goal, while a higher score indicates meaningful forward movement. EP may not make sense in scenarios where a sudden obstacle appears or other unexpected events occur; question EP if the camera view contradicts it.
    - Comfort: A lower score suggests uncomfortable driving behavior, such as sudden braking, jerky motion, or harsh acceleration/deceleration.
All visible traffic signs and road markings follow U.S. traffic rules, where vehicles drive on the right-hand side of the road. Yellow lines separate traffic moving in opposite directions, while white lines separate lanes traveling in the same direction.
Determine, for each scenario, which metrics should be used to evaluate the driving behavior.
For example, in a lane nudging scenario, evaluating with LK and DDC might not make sense due to temporarily taking an incorrect lane.
When providing your analysis, exclude the following:
    - Unrelated objects
    - Hypothetical scenarios or actions
    - Street or business names
    - Vehicle makes or models
    
User: 
This is a short clip (four frames in temporal order) of a driving scenario.
<video>
Each frame contains a frontal camera view on the left and a Bird's-Eye View (BEV) image on the right.
Each frame shows the vehicle's future 4-second driving trajectory.
In the frontal view image, the trajectory is overlaid in a cyan dotted line.
In the BEV image, the EGO vehicle and its trajectory are shown in RED, while bounding boxes of OTHER AGENTS are displayed in BLUE.
On the BEV, lane boundaries are drawn in solid black lines, and lane CENTERLINES are shown as BLUE arrows that also indicate the normal driving direction of each lane.

The seven criteria (fixed order below):
- LK - Lane Keeping
- DDC - Driving Direction Compliance
- DAC - Drivable Area Compliance
- TTC - Time to Collision
- NC - No Collision
- EP - Ego Progress
- COMFORT - Comfort

Decide whether LK, DDC, DAC, TTC, NC, EP, and COMFORT should each be leveraged to evaluate driving performance in this scenario. Do not judge whether the driving violates a criterion; only whether that criterion is useful to evaluate performance here.

Reply with exactly one line after the prompt below: seven labels in this exact order - LK, then DDC, then DAC, then TTC, then NC, then EP, then COMFORT - separated by commas and spaces. Each label must be exactly Useful or Questionable.

Example format (values are illustrative): Useful, Questionable, Useful, Useful, Useful, Useful, Useful

Requirements:
- Exactly seven labels, in the fixed order above
- Only the words Useful or Questionable; no other text on that line

Your answer for criterion usefulness for driving evaluation for LK, DDC, DAC, TTC, NC, EP, and COMFORT is:

\end{lstlisting}

\subsection{Detail on the non-vlm baseline} 
\label{subsec:non_vlm_detail}
To ablate the use of a VLM model, we present a simple visual classifier baseline as the replacement model backbone. We use Qwen3-VL-Embedding-2B~\cite{qwen3vlembedding} to extract visual embeddings from the same input images and we train a simple linear logistic classifier on top of it.
We train two models, Non-VLM and Non-VLM-Agent.
Non-VLM is a direct judgement classifier trained with the human reasonable labels $\mathbf{y}$.
Non-VLM-Agent makes use of PDM in a very similar way as~\method{}. The model is trained for binary classification for rule-invocation using the same ground-truth rule invocation labels $g_i^m$. A classification threshold is selected based on the training set and used to infer rule-invocations on the held out test set. The predicted rule invocations are aggregated following Equation~\ref{eqn:drivejudge}.
We report scores of the two ablation models in Table~\ref{tab:ablation_main} showing some gain compared to the zero-shot VLM-DirectScore approach but still underperforming due to the limited contextual understanding of the non-VLM model. We don't see a significant difference between adding or not adding rule invocation showing that the benefit of using rule invocation requires the base model to have enough underlying capability to use it well.

\subsection{VLM direct scoring baseline}
\begin{lstlisting}[caption={VLM Direct Score Prediction Prompt}]
System: You are an expert AI driving instructor and analyst. Your task is to evaluate whether the driving behavior shown in the video is reasonable and does not violate any best driving practices.
Focus on the following aspects when forming your judgment:
- Lane Keeping: Is the vehicle staying appropriately within its lane, accounting for legitimate maneuvers like lane changes, overtakes, or navigating construction zones?
- Driving Direction: Is the vehicle following the intended direction of travel on each road?
- Drivable Area: Is the vehicle remaining within the designated drivable area?
- Safety around other agents: Is the trajectory safe with respect to other vehicles, pedestrians, and obstacles?
- Ego Progress: Is the vehicle making reasonable forward progress toward its goal, given the road conditions and any obstacles?
- Comfort: Is the driving behavior smooth, without harsh braking, jerky motion, or sudden acceleration?
All visible traffic signs and road markings follow U.S. traffic rules, where vehicles drive on the right-hand side of the road. Yellow lines separate traffic moving in opposite directions, while white lines separate lanes traveling in the same direction.
When providing your analysis, exclude the following:
- Unrelated objects
- Hypothetical scenarios or actions
- Street or business names
- Vehicle makes or models

User: <video> 
These are video frames from a driving video.
Each frame contains a frontal camera view on the left and a Bird's-Eye View (BEV) image on the right.
Each frame shows the vehicle's future 4-second driving trajectory.
In the frontal view image, the trajectory is overlaid in a cyan dotted line.
In the BEV image, the EGO vehicle and its trajectory are shown in RED, while bounding boxes of OTHER AGENTS are displayed in BLUE.
On the BEV, lane boundaries are drawn in solid black lines, and lane CENTERLINES are shown as BLUE arrows that also indicate the normal driving direction of each lane.

Describe the driving scenario shown in the video, then provide a single score between 0 and 1 that reflects how reasonable and safe the ego vehicle's driving behavior is.
A score of 1.0 means the driving is completely reasonable and follows all best driving practices.
A score of 0.0 means the driving is completely unreasonable or unsafe.

All responses must be in English only.
Do not mention anything you are not certain about - avoid statements with 'maybe', 'might', or 'possibly'.
The description should be concise and to the point.

**CRITICAL: You MUST format your response as a valid JSON object with this EXACT structure:**

```json
{
  "context": "Brief description of the driving scenario and key observations",
  "reasoning": "Concise explanation of why this score was assigned"
  "score": 0.85,
}
```

**Requirements:**
- "score" must be a single float in the range [0.0, 1.0]
- "reasoning" should clearly justify the score, referencing specific behaviors observed
- Return ONLY the JSON object, no additional text before or after

\end{lstlisting}

\section{Data}
\subsection{Thresholds per metric for violation when creating the groundtruth gating}
\label{subsec:thresholds}
As mentioned in Section~\ref{sec:SFT}, we construct ground-truth rule invocation labels using the human annotations in~\method{} dataset along with the computed EPDMS metrics. We do this by disabling tool invocations that would result in low scores for samples that are labeled as reasonable driving.
The violation thresholds $\tau^m$ per EPDMS metric defines the lowest score value that can be tolerated below which their invocation is disabled in reasonable clips. We select them to maximize the alignment between the resulting score with human judgement in our dataset.

To do this we sweep over a range of different settings for $\tau^m$ within reasonable bounds. For each threshold setting, we use the implied ground truth tool invocations $g_i^m$ to compute the final scores $S_i$. We then select the threshold setting that leads to the predicted scores with the highest accuracy with respect to human annotator driving quality label:
\begin{align}
\text{arg}\max_{\tau^m} \mathrm{ACC_{DQC}}(S(g_i^m(\tau_m)), \mathbf{y})
\end{align}

On our dataset we found the threshold setting achieving the highest alignment with human judgement ($\mathrm{ACC_{DQC}}=93.80$) to be:
\begin{table}[h!]
\centering
\begin{tabular}{ccccccc}
\toprule
LK & NC & DAC & DDC & TTC & COMFORT & EP \\
\midrule
0.7 & 0.9 & 0.4 & 0.7 & 0.9 & 0.6 & 0.5 \\
\bottomrule
\end{tabular}
\end{table}

\begin{figure}[htbp]
  \centering

  \begin{subfigure}[b]{0.48\textwidth}
    \centering
    \clipbox{0 0 0.4\width{} 0}{\includegraphics[width=1.6\textwidth]{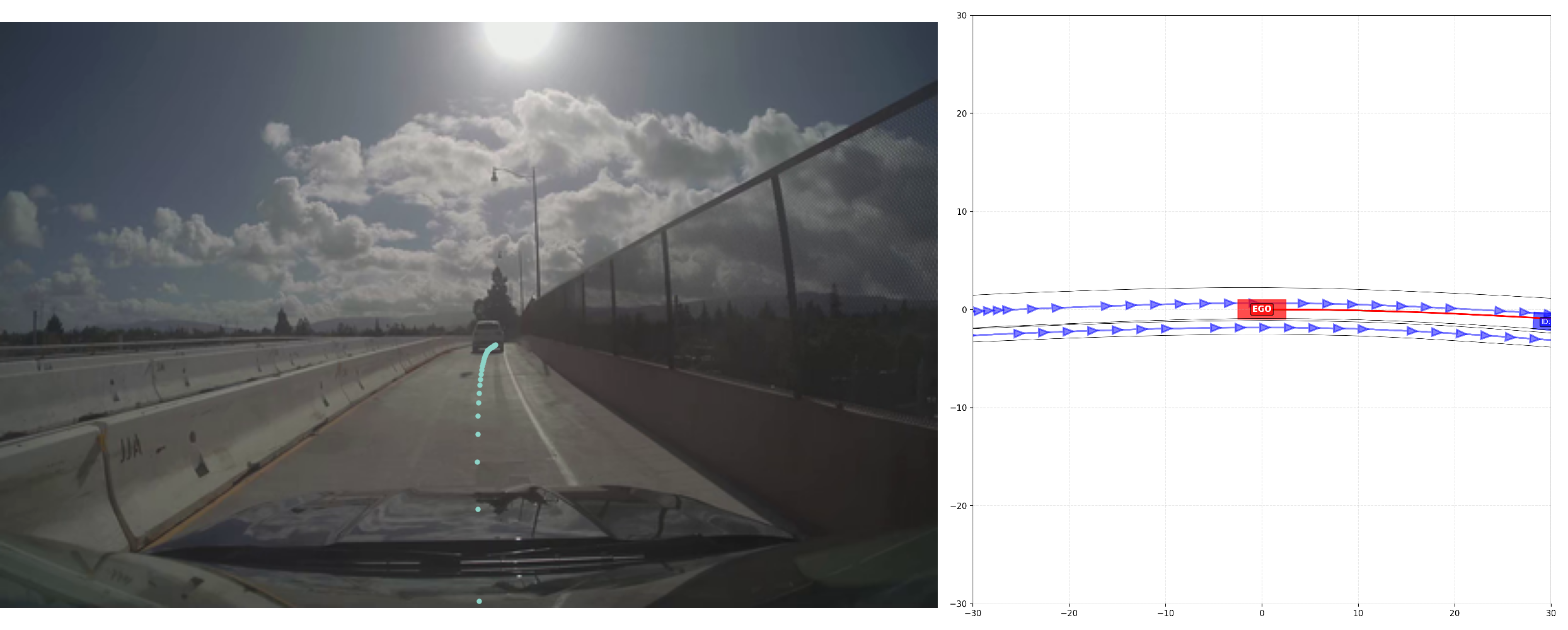}}
    \caption{Reasonable: Ego drifts right in lane to keep buffer from barrier on left.}
  \end{subfigure}
  \hfill
  \begin{subfigure}[b]{0.48\textwidth}
    \centering
    \clipbox{0 0 0.4\width{} 0}{\includegraphics[width=1.6\textwidth]{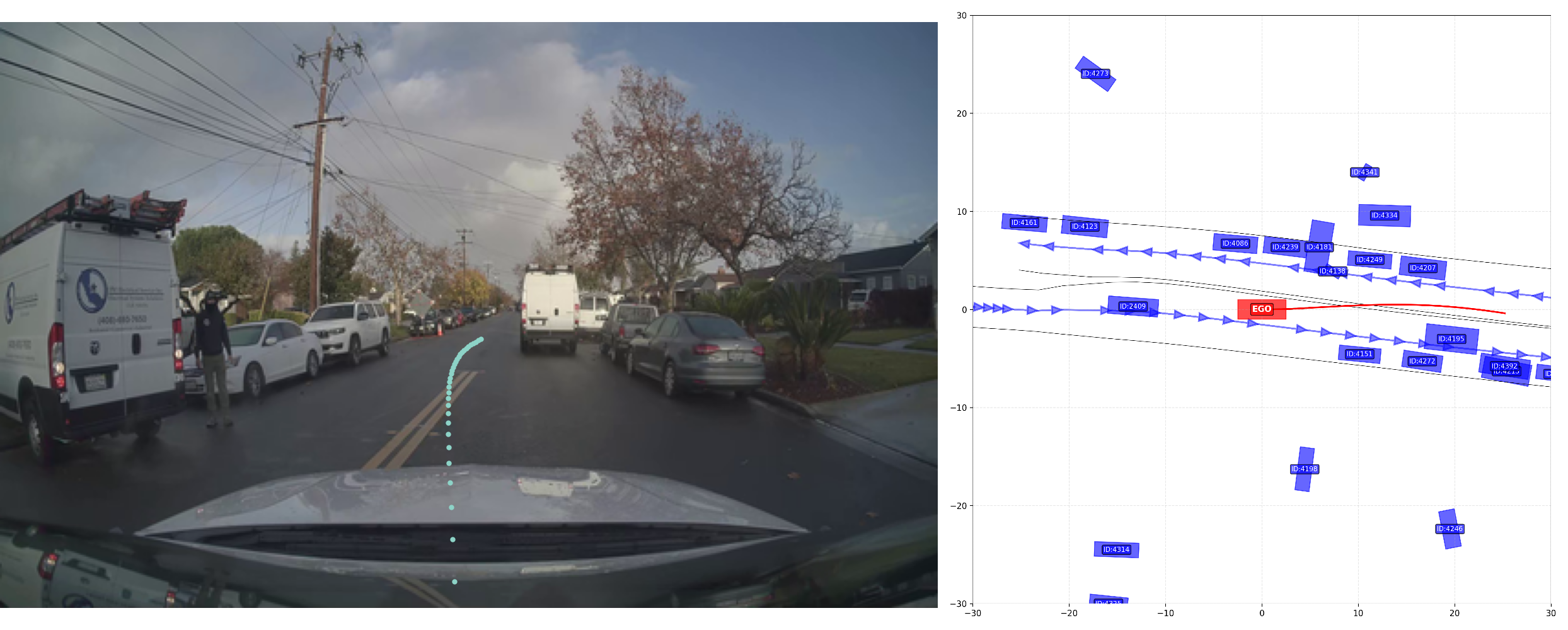}}
    \caption{Reasonable: Ego nudges left across double yellow lane line due to stopped truck in front.}
  \end{subfigure}

  \vspace{1em}

  \begin{subfigure}[b]{0.48\textwidth}
    \centering
    \clipbox{0 0 0.4\width{} 0}{\includegraphics[width=1.6\textwidth]{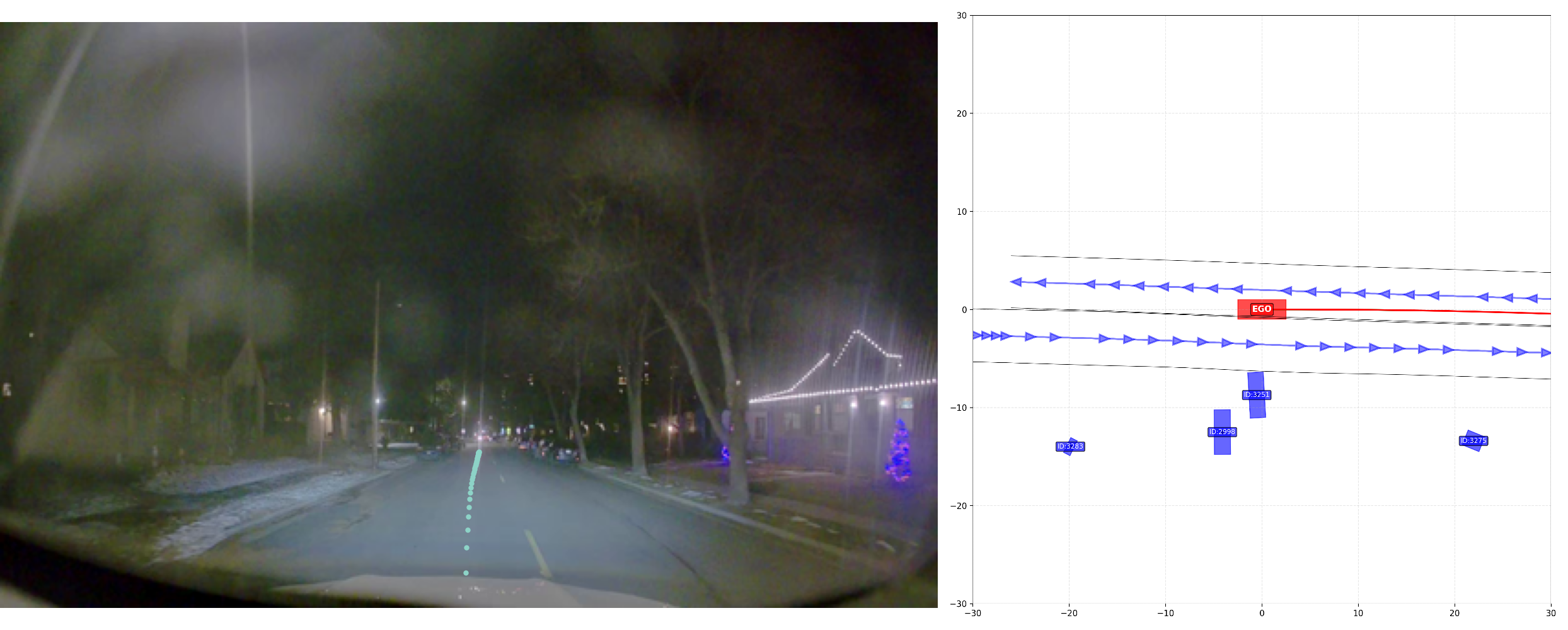}}
    \caption{Unreasonable: Ego is driving in opposite direction lane too long.}
  \end{subfigure}
  \hfill
  \begin{subfigure}[b]{0.48\textwidth}
    \centering
    \clipbox{0 0 0.4\width{} 0}{\includegraphics[width=1.6\textwidth]{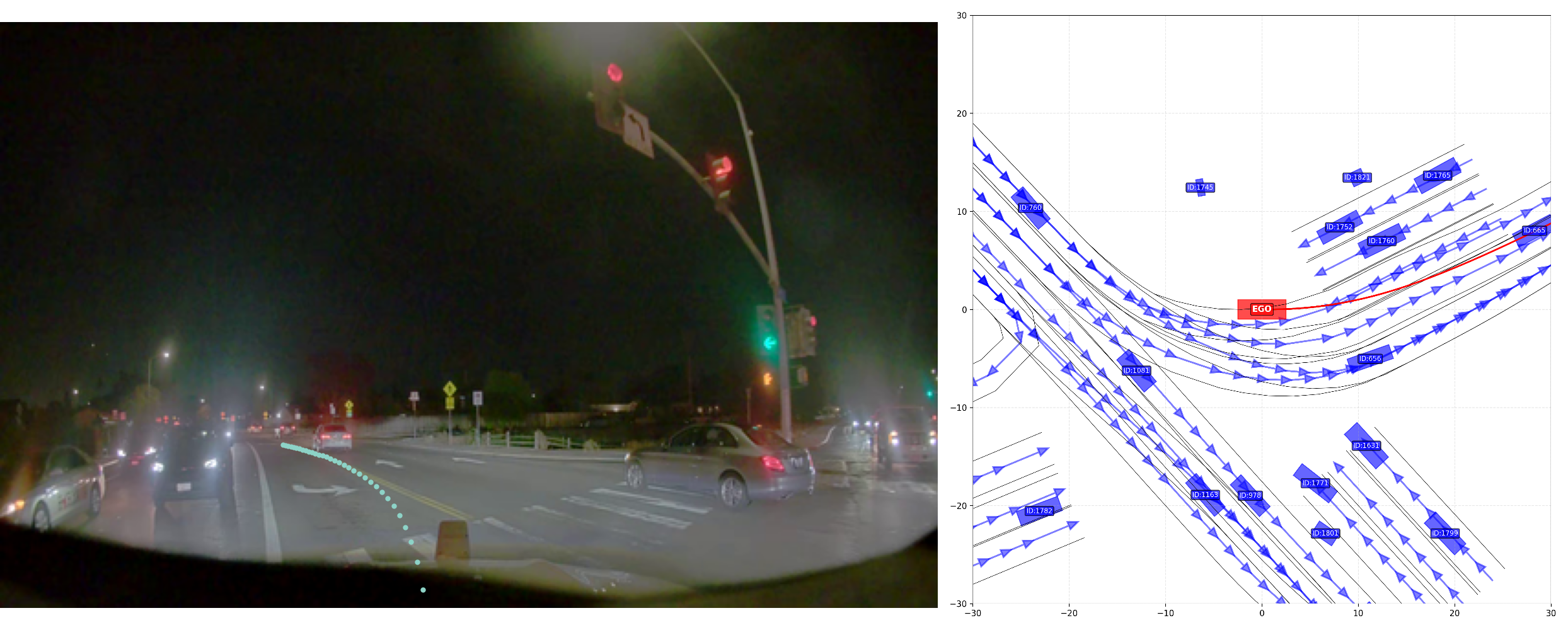}}
    \caption{Unreasonable: Ego is cutting corner on left turn and drives through opposite lane.}
  \end{subfigure}

  \vspace{1em}

  \begin{subfigure}[b]{0.48\textwidth}
    \centering
    \clipbox{0 0 0.4\width{} 0}{\includegraphics[width=1.6\textwidth]{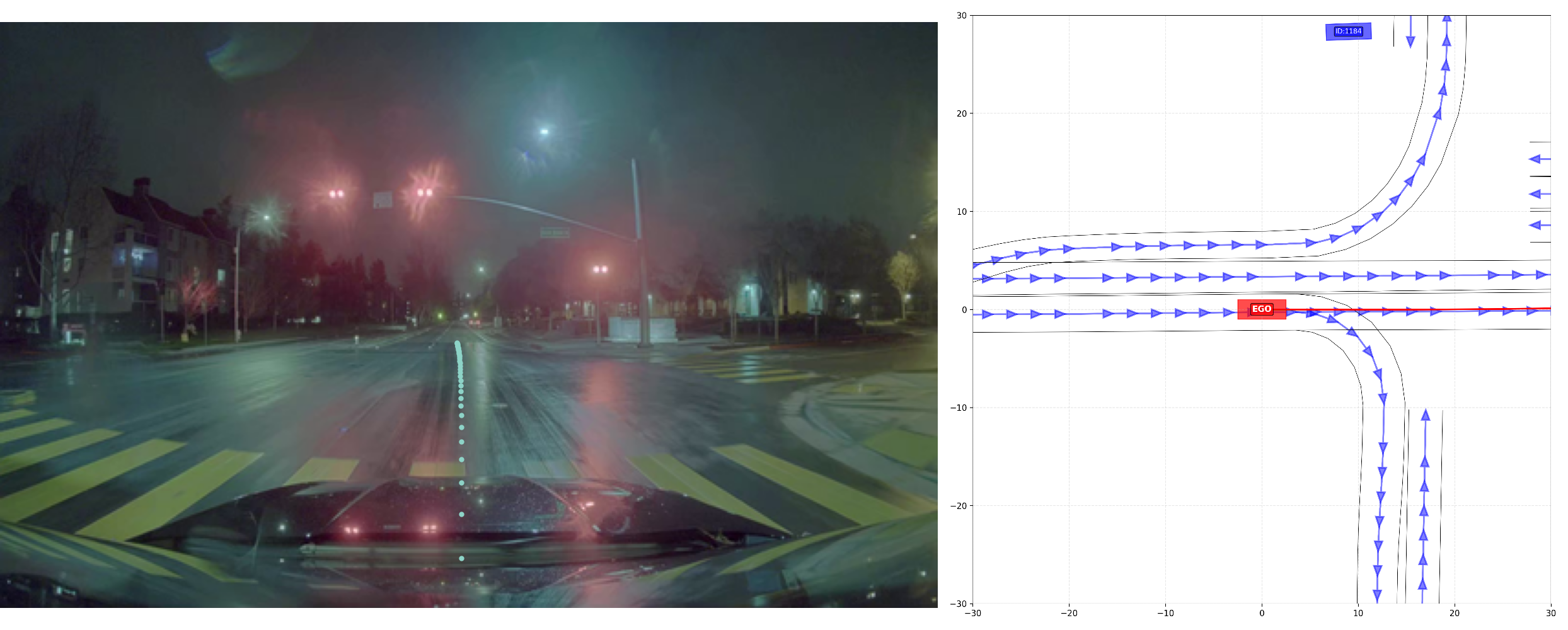}}
    \caption{Unreasonable: Ego runs a red light driving straight through the intersection.}
  \end{subfigure}
  \hfill
  \begin{subfigure}[b]{0.48\textwidth}
    \centering
    \clipbox{0 0 0.4\width{} 0}{\includegraphics[width=1.6\textwidth]{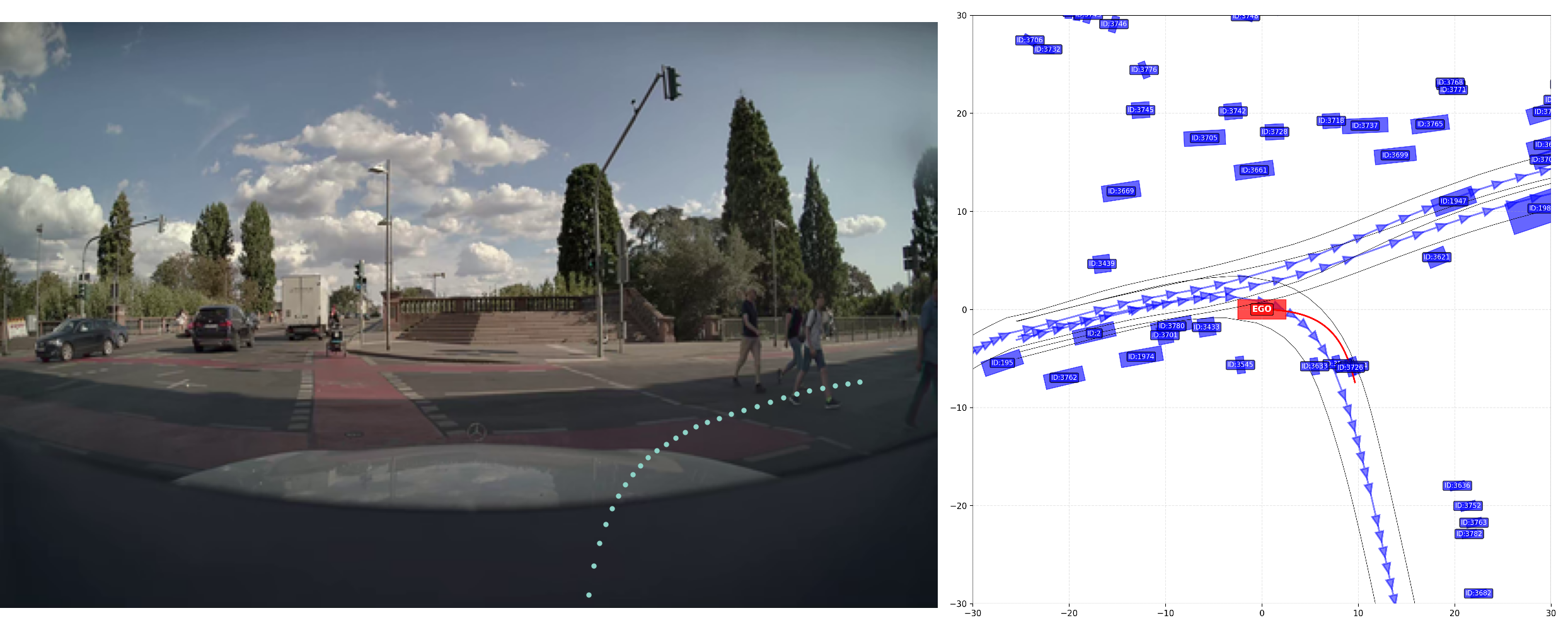}}
    \caption{Unreasonable: Ego is creeping too close to pedestrians on a right turn.}
  \end{subfigure}
  \caption{Representative examples from our collected annotation labels.}
  \label{fig:annotation_examples}

\end{figure}

\subsection{Examples from the annotations}
We show some representative examples of collected annotations from our~\method{} dataset in Figure~\ref{fig:annotation_examples}.

\label{subsec:navsim_filtering}
\begin{figure}[ht!]
    \centering

    \begin{subfigure}[t]{\linewidth}
        \centering
        \includegraphics[width=\linewidth]{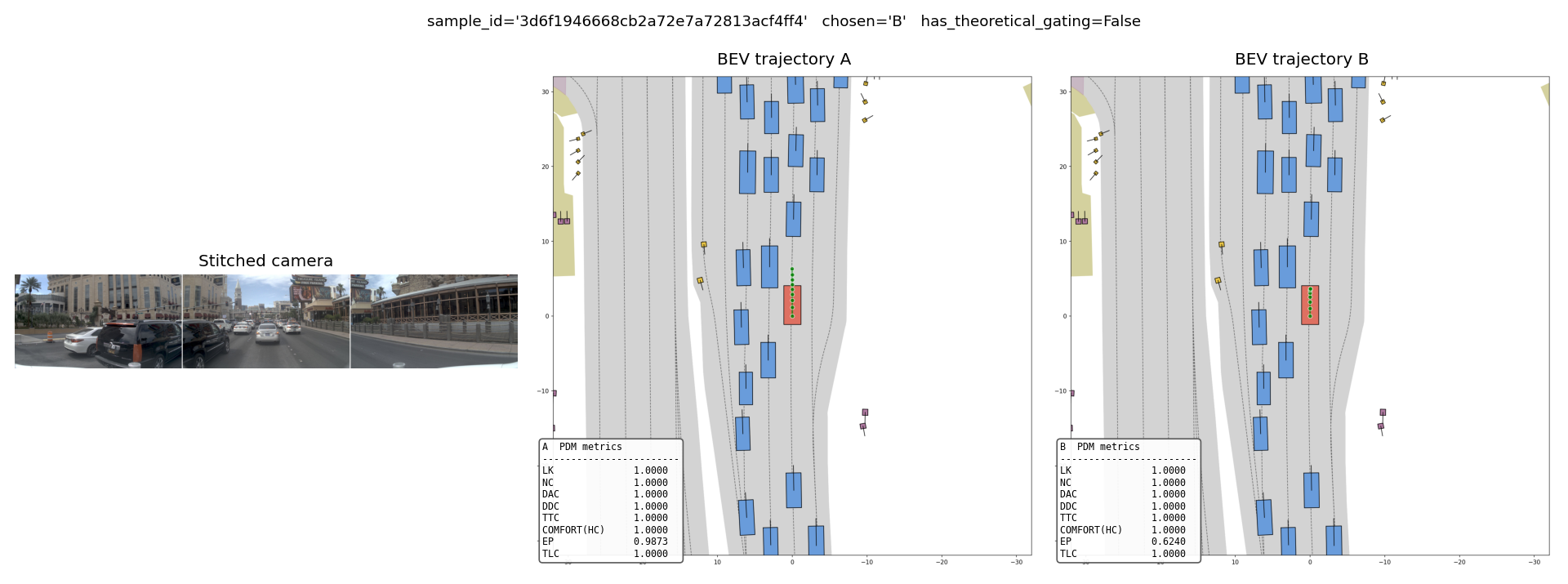}
        \label{fig:a}
    \end{subfigure}

    \vspace{4pt}

    \begin{subfigure}[t]{\linewidth}
        \centering
        \includegraphics[width=\linewidth]{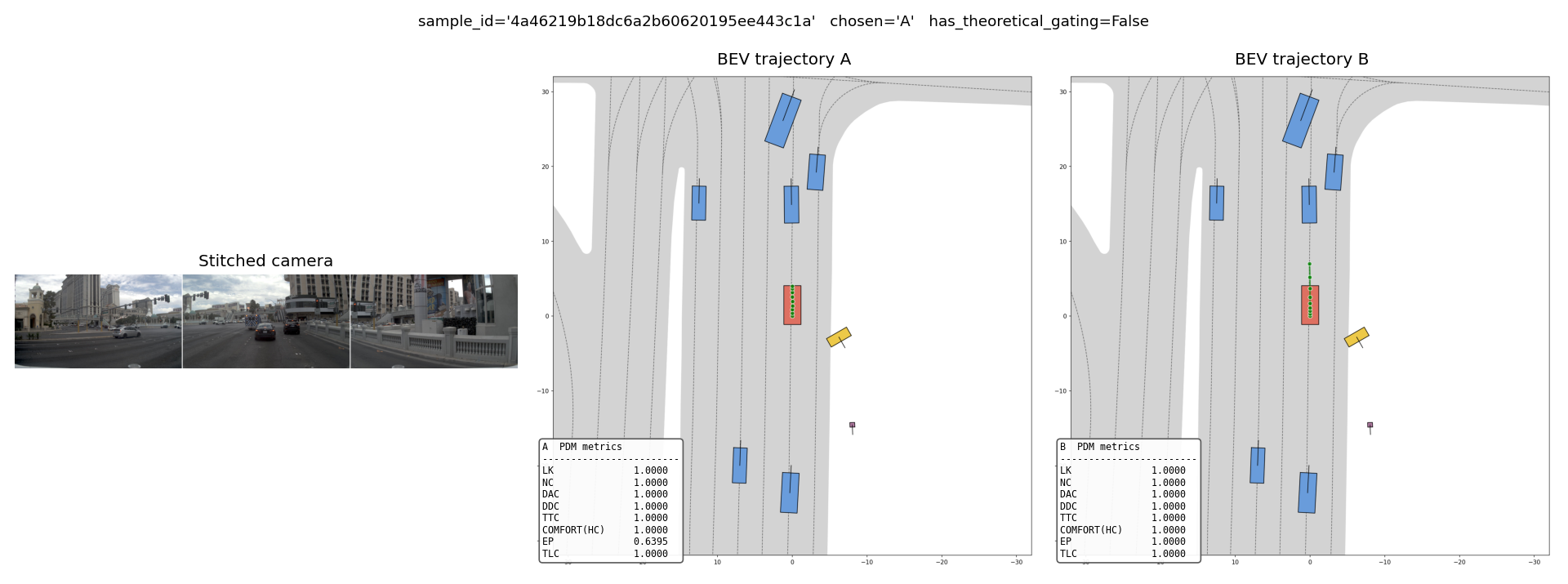}
        \label{fig:b}
    \end{subfigure}

    \vspace{4pt}

    \begin{subfigure}[t]{\linewidth}
        \centering
        \includegraphics[width=\linewidth]{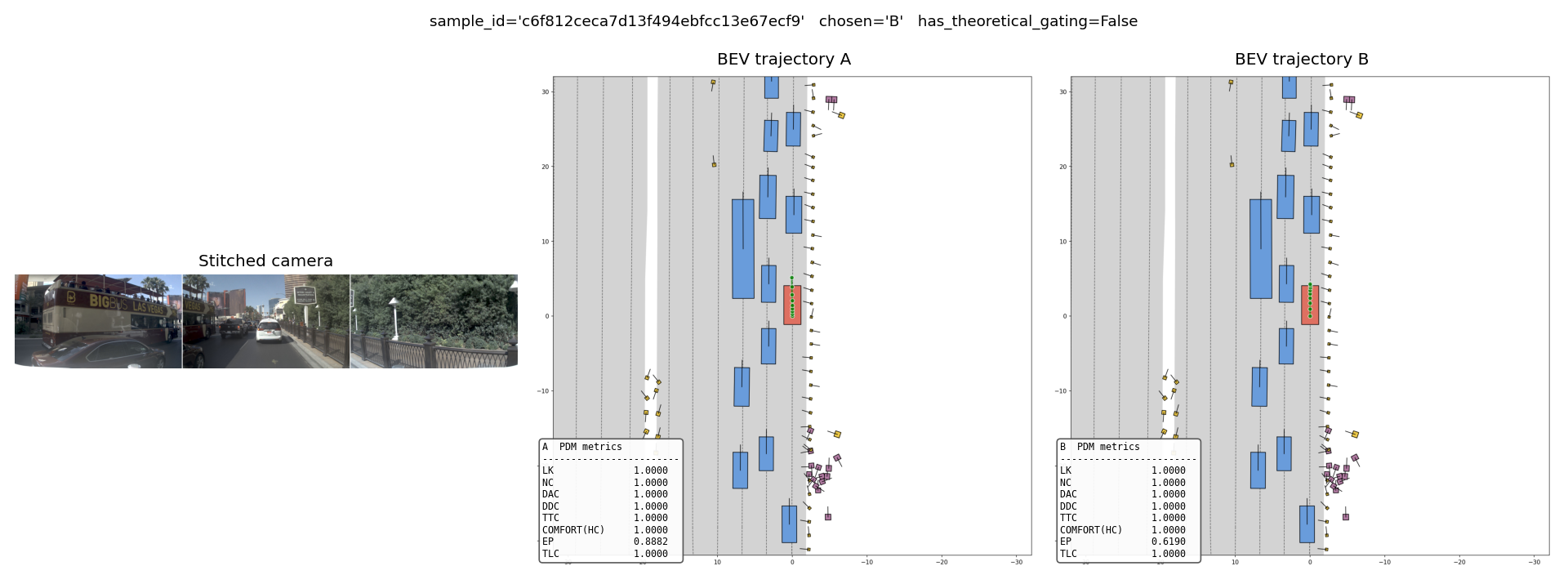}
        \label{fig:c}
    \end{subfigure}

    \caption{\textbf{Ambiguous preference samples removed from DriveCritic~\cite{song2025drivecritic}.} 
A common pattern in the original DriveCritic dataset involves trajectory pairs without a clear human preference. In this example, both trajectories are safe and valid, differing only in minor progress behavior (e.g., slowly advancing versus remaining stationary), making the preference inherently ambiguous. We remove such samples to construct a cleaner trajectory preference benchmark.}
    \label{fig:drivecritic_filtered_out}
    \vspace{-20pt}
\end{figure}
\subsection{Filtering the DriveCritic~\cite{song2025drivecritic} Trajectory Selection Dataset}
\label{subsec:navsim_filtering}
The original DriveCritic~\cite{song2025drivecritic} trajectory preference dataset contains a substantial number of samples where the preference signal is inherently ambiguous, as shown in Fig~\ref{fig:drivecritic_filtered_out}. A common pattern we observe is that both candidate trajectories are safe, valid, and largely equivalent in driving quality, differing only in minor behavioral details such as slightly different progress profiles (e.g., slowly creeping forward versus remaining stationary). In such cases, assigning a strict preference between the two trajectories is often subjective and does not reflect a meaningful difference in driving quality.

For evaluation, these ambiguous samples weaken the reliability of the benchmark itself. If both trajectories are equally reasonable under the driving context, forcing a binary preference label introduces noise into the evaluation target and makes metric comparison less informative. A strong driving metric should distinguish meaningful quality differences, rather than arbitrary preferences between equally valid behaviors.

For training, these samples also introduce undesirable supervision noise. Preference optimization methods assume that the preference label reflects a consistent quality ordering between candidates. When this assumption does not hold, the optimization objective may encourage the model to learn unstable or non-generalizable preference patterns.

To improve benchmark quality, we manually filter out these ambiguous trajectory pairs and retain only samples where one trajectory exhibits a clear and meaningful advantage under human judgment, such as stronger safety, better progress, or more appropriate rule compliance under the scene context. After filtering, the validation split is reduced from 1,166 to 862 samples, and the training split is reduced from 4,564 to 3,597 samples. The resulting dataset provides a cleaner and more reliable benchmark for trajectory preference evaluation and preference-based post-training.
\section{Method}
\subsection{Full Scoring Formula}
\begin{align}
\label{eqn:full_formula}
    \mathcal{S}_i =
    \left(
    \prod_{m \in \mathcal{M}_{\mathrm{pen}}}
    (\hat{g}_i^m s_i^m + (1-\hat{g}_i^m))
    \right)
    \cdot
    \frac{
    \sum_{m \in \mathcal{M}_{\mathrm{avg}}} \hat{g}_i^m w^m s_i^m
    }{
    \sum_{m \in \mathcal{M}_{\mathrm{avg}}} \hat{g}_i^m w^m
    }.
\end{align}

Concretely, the full driving score slightly differs from the simplified form shown in Eqn.~\ref{eqn:drivejudge_score}. We present the complete formulation in Eqn.~\ref{eqn:full_formula}. Here, $\mathcal{M}{\mathrm{pen}}$ denotes hard safety metrics aggregated multiplicatively (including NC, DAC, and DDC), while $\mathcal{M}{\mathrm{avg}}$ represents the remaining soft metrics combined via weighted averaging. The weight $w^m$ specifies the relative contribution of each invoked sub-metric, following~\cite{dauner2024navsim,cao2025pseudo}.

\section{Experiments}
\begin{figure}
    \centering
    \includegraphics[width=.7\linewidth]{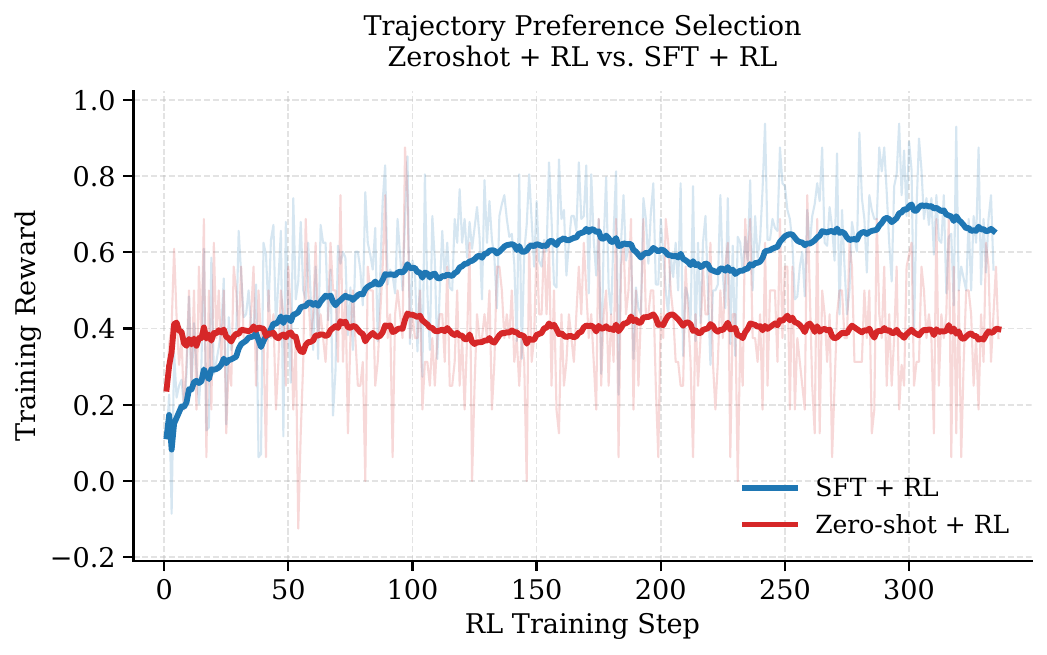}
    \caption{\textbf{RL training reward comparison from different initialization.} 
Starting from an SFT-initialized model leads to steady reward improvement, while zero-shot initialization quickly plateaus with minimal gain. This highlights the importance of SFT as a critical warm start.}
    \label{fig:reward_comparison}
\end{figure}

\subsection{Training Reward Curve Comparison between Zero-shot+RL and SFT+RL}

In the main paper, we compare Zero-shot+RL and SFT+RL through validation accuracy on the trajectory preference benchmark, showing that reinforcement learning on top of supervised fine-tuning leads to substantially better downstream performance. Here, we provide the corresponding training reward curves to further illustrate the optimization dynamics behind this difference.

Figure~\ref{fig:reward_comparison} shows the reward progression during RL training under identical optimization settings, with the only difference being the initialization: either starting directly from the zero-shot model or from the SFT-initialized checkpoint. The reward is defined by Eq.~\ref{eqn:reward}, measuring whether the model’s trajectory ranking aligns with the human preference label.

A clear difference emerges in training behavior. Starting from the SFT-initialized model leads to stable and consistent reward improvement throughout RL optimization, indicating effective exploration and continual policy refinement. In contrast, starting from the zero-shot model quickly plateaus at a much lower reward level, with little meaningful improvement over training.

This result highlights the importance of SFT as a warm-start stage before RL. While RL is effective at refining preference alignment, it relies on an initial policy with sufficient task understanding to explore meaningful rule invocation strategies. Without this initialization, the model struggles to discover useful behaviors, resulting in weak reward improvement and limited downstream gains.

\subsection{Training curves}
\begin{figure}[ht!]
    \centering

    \begin{subfigure}[t]{0.49\linewidth}
        \centering
        \includegraphics[width=\linewidth]{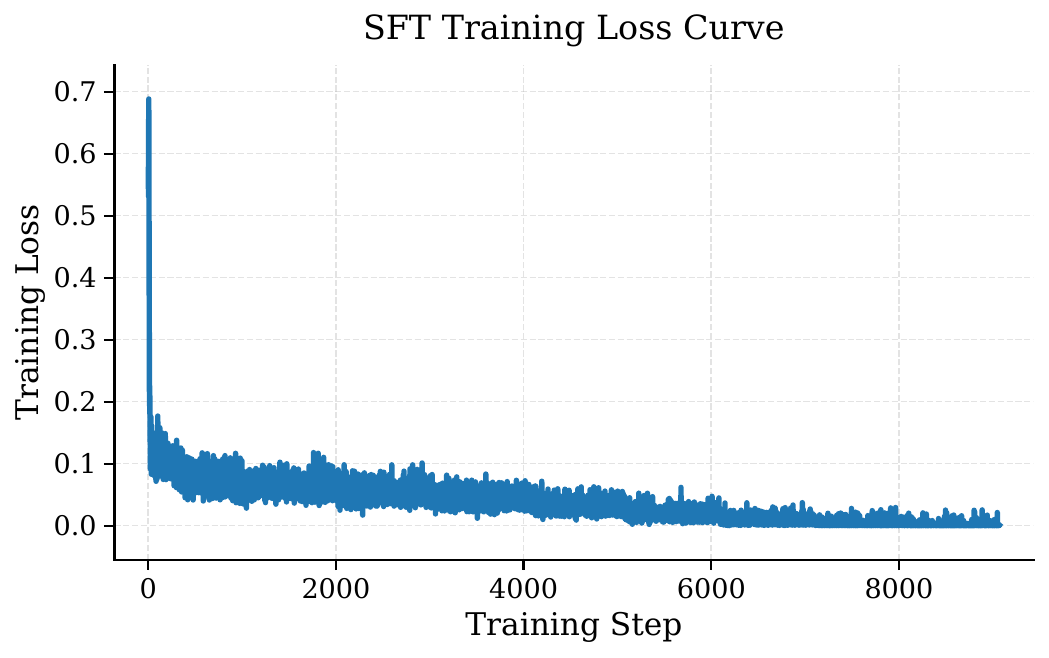}
        \caption{SFT training loss curve.}
        \label{fig:sft_loss_curve}
    \end{subfigure}
    \hfill
    \begin{subfigure}[t]{0.49\linewidth}
        \centering
        \includegraphics[width=\linewidth]{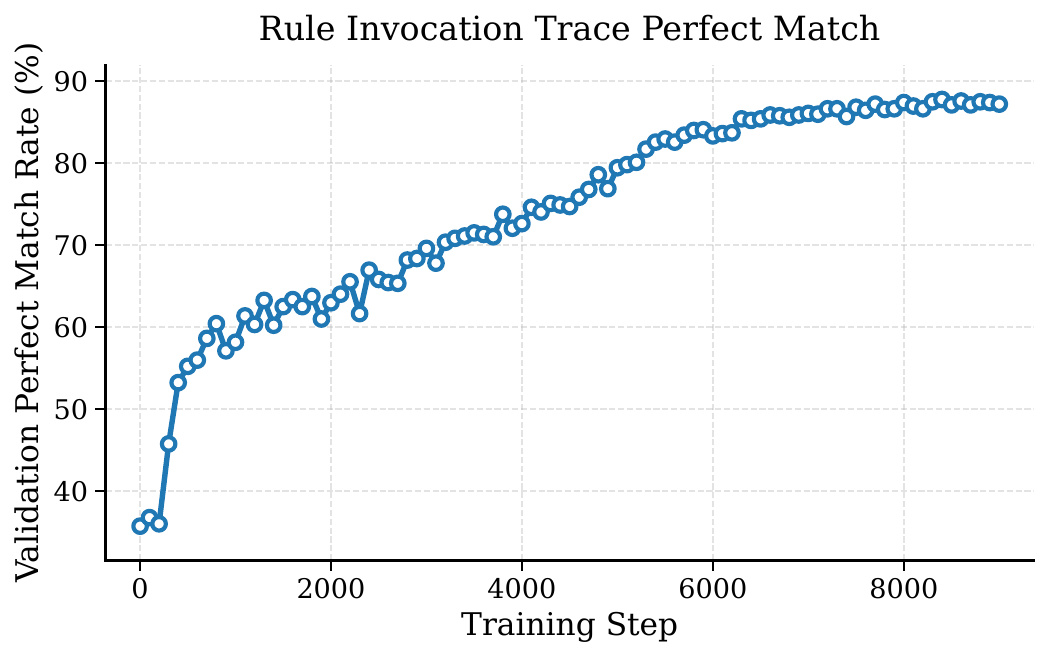}
        \caption{Validation perfect-match rate.}
        \label{fig:sft_perfect_match}
    \end{subfigure}

    \caption{\textbf{SFT training dynamics.} Left: training loss over optimization steps. Right: validation perfect-match rate measuring exact alignment between predicted rule invocation traces and ground-truth traces.}
    \label{fig:training_curves}
\end{figure}

Figure~\ref{fig:training_curves} presents the supervised fine-tuning (SFT) training dynamics of \method{}-8B. In addition to downstream benchmark performance reported in the main paper, these curves provide a direct view into optimization behavior and the model’s ability to learn the rule invocation task.Figure~\ref{fig:sft_loss_curve} shows the training loss over optimization steps. The loss decreases steadily throughout training, indicating stable optimization and effective fitting to the curated supervision signals. The rapid initial drop suggests that the model quickly adapts to the task-specific output format and rule invocation patterns, while the slower later-stage decline reflects gradual refinement of decision boundaries.Figure~\ref{fig:sft_perfect_match} shows the validation perfect-match rate, which measures the percentage of validation samples where the predicted rule invocation trace exactly matches the ground-truth rule invocation trace across all evaluation dimensions. This is a stricter metric than per-rule accuracy, as a sample is counted as correct only if the entire rule invocation sequence is predicted perfectly. We observe consistent improvement throughout training, eventually reaching close to $90\%$, demonstrating that \method{}-8B learns to reliably recover the intended context-aware rule selection behavior.Together, these training curves show that supervised fine-tuning effectively improves the rule invocation capability of \method{}, providing a strong initialization for the subsequent reinforcement learning stage.

\subsection{Hyperparameters for RL}

Table~\ref{tab:rl_hyperparams} presents the ablation of key hyperparameters used in the reinforcement learning (RL) stage for \method{}-2B, including the KL loss coefficient, number of rollouts, and batch size. We report the final trajectory preference selection accuracy under each setting.

We observe that the KL loss coefficient has a noticeable impact on performance. A relatively large coefficient ($0.1$) leads to worse performance, likely because the optimization is overly constrained toward the reference policy, limiting effective policy improvement. A very small coefficient ($0.001$) improves flexibility but remains slightly suboptimal. In practice, a moderate coefficient ($0.01$) achieves the best trade-off between preserving policy stability and allowing reward-driven exploration.

The number of rollouts also affects performance. Reducing the rollout size from $8$ to $4$ leads to a noticeable drop in accuracy (80.63$\rightarrow$77.63), suggesting that sufficient rollout diversity is important for stable relative advantage estimation under GRPO.

Finally, batch size influences optimization stability. Reducing the batch size from $32$ to $16$ also degrades performance (80.63$\rightarrow$78.31), likely due to noisier gradient estimates and less stable policy updates.

Based on these results, we adopt a KL loss coefficient of $0.01$, rollout size of $8$, and batch size of $32$ as the default configuration for all RL experiments in this work.

\begin{table}[ht!]
\centering
\caption{\textbf{Reinforcement learning hyperparameters and performance} from \method{}-2B.}
\label{tab:rl_hyperparams}
\small
\begin{tabular}{c c c | c}
\toprule
KL Loss Coefficient & \# Rollouts & Batch Size & Accuracy (\%) \\
\midrule
0.001 & 8 & 32 & 79.35 \\
0.1 & 8 & 32 & 78.75 \\
0.01 & 4 & 32 & 77.63 \\
0.01 & 8 & 16 & 78.31 \\
\rowcolor{lgreen}
0.01 & 8 & 32 & 80.63 \\
\bottomrule
\end{tabular}
\end{table}

\subsection{BEV VLM input ablation for zero-shot evaluation}

Table~\ref{tab:bev_ablation} studies the effect of providing bird’s-eye-view (BEV) inputs during zero-shot evaluation. Specifically, we ablate whether to render the BEV representation and provide it to the VLM together with the frontal camera image as multi-modal input.

\begin{table}[h!]
\centering
\caption{\textbf{BEV ablation for zero-shot evaluation.} 
Effect of providing bird’s-eye-view (BEV) inputs to the VLM during zero-shot driving quality classification.}
\label{tab:bev_ablation}
\small
\begin{tabular}{c c}
\toprule
BEV Input & AUC \\
\midrule
\xmark & 52.39 \\
\rowcolor{lgreen}
\cmark & 56.13 \\
\bottomrule
\end{tabular}
\end{table}

The BEV representation provides structured spatial information, including lane topology, surrounding agents, and geometric layout, which complements the egocentric camera view. This additional spatial context helps the VLM better reason about scene structure and driving constraints.

As shown in Table~\ref{tab:bev_ablation}, incorporating BEV improves zero-shot performance from 52.39 to 56.13 AUC. This result highlights the importance of explicit spatial representations for improving VLM-based driving evaluation, particularly in zero-shot settings where no task-specific adaptation is available.


\end{document}